\documentclass[11pt]{article}
\usepackage[final]{acl}
\usepackage{times}
\usepackage{latexsym}
\usepackage[T1]{fontenc}
\usepackage[utf8]{inputenc}
\usepackage{microtype}
\usepackage{inconsolata}

\usepackage{listings}
\usepackage{xcolor}
\usepackage{colortbl}
\usepackage[most]{tcolorbox}
\usepackage{caption}

\usepackage{graphicx}
\everymath{\textstyle}

\title{\textsc{Vibe-Bench}: Evaluating Personalized Large Language Models \linebreak When Profiles Don't Mean Preferences}

\author{
\textbf{Yiwen Jiang} $^{1}$ \quad
\textbf{Yang Deng} $^{2}$ \quad
\textbf{Stephanie Fong} $^{1}$ \quad
\textbf{Zimu Wang} $^{3}$ \quad \\
\textbf{Yaling Shen} $^{1}$ \quad
\textbf{Wei Feng} $^{1}$ \quad
\textbf{Hongxi Yang} $^{1}$ \quad
\textbf{Xiangyu Zhao} $^{1}$ \\
\textbf{Zhongxing Xu} $^{1}$ \quad
\textbf{Deval Mehta} $^{1,4}$ \quad
\textbf{Xuelian Cheng} $^{1}$ \quad
\textbf{Zongyuan Ge} $^{1,\dagger}$ \\
\textsuperscript{1}Monash University \quad
\textsuperscript{2}Singapore Management University \\
\textsuperscript{3}University of Liverpool \quad
\textsuperscript{4}RMIT University
\\
\texttt{\{yiwen.jiang, zongyuan.ge\}@monash.edu}
}

\usepackage{amsmath}
\usepackage{enumitem}
\usepackage{makecell}
\usepackage{multirow}

\newcommand{\iconriasec}{\raisebox{-0.15em}{\includegraphics[height=0.9em]{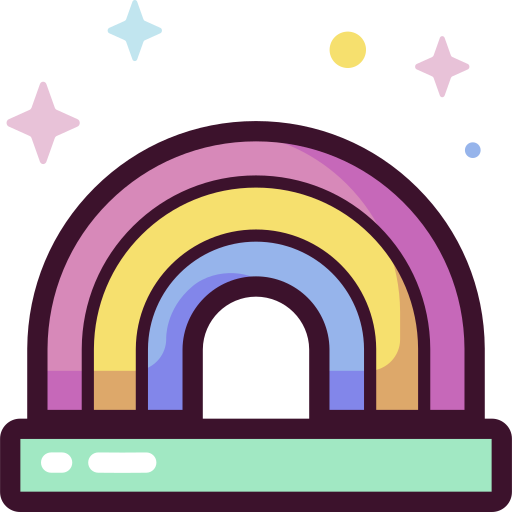}}}
\newcommand{\iconbigfiveup}{\raisebox{-0.15em}{\includegraphics[height=0.9em]{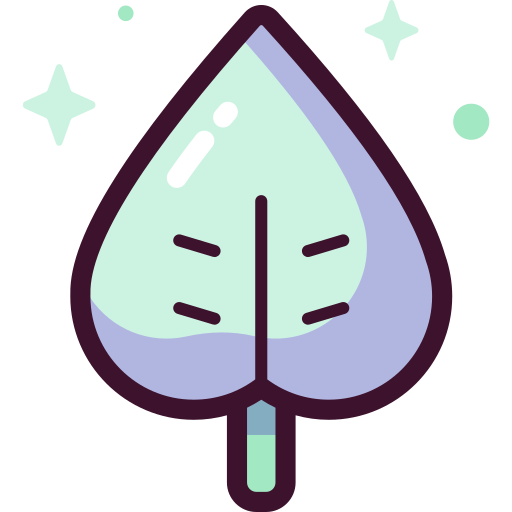}}}
\newcommand{\iconbigfivedown}{\raisebox{-0.15em}{\includegraphics[height=0.9em]{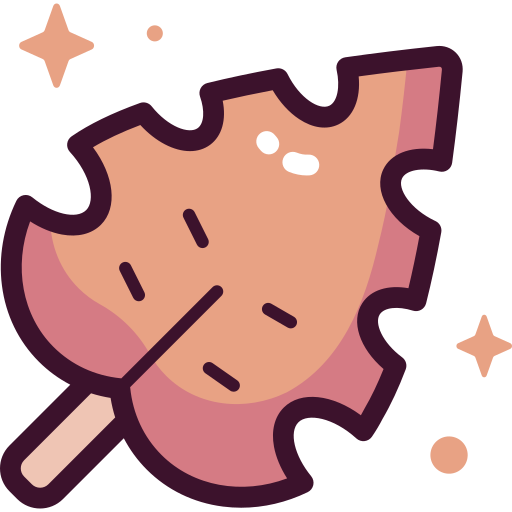}}}
\newcommand{\iconqwen}{\raisebox{-0.15em}{\includegraphics[height=0.95em]{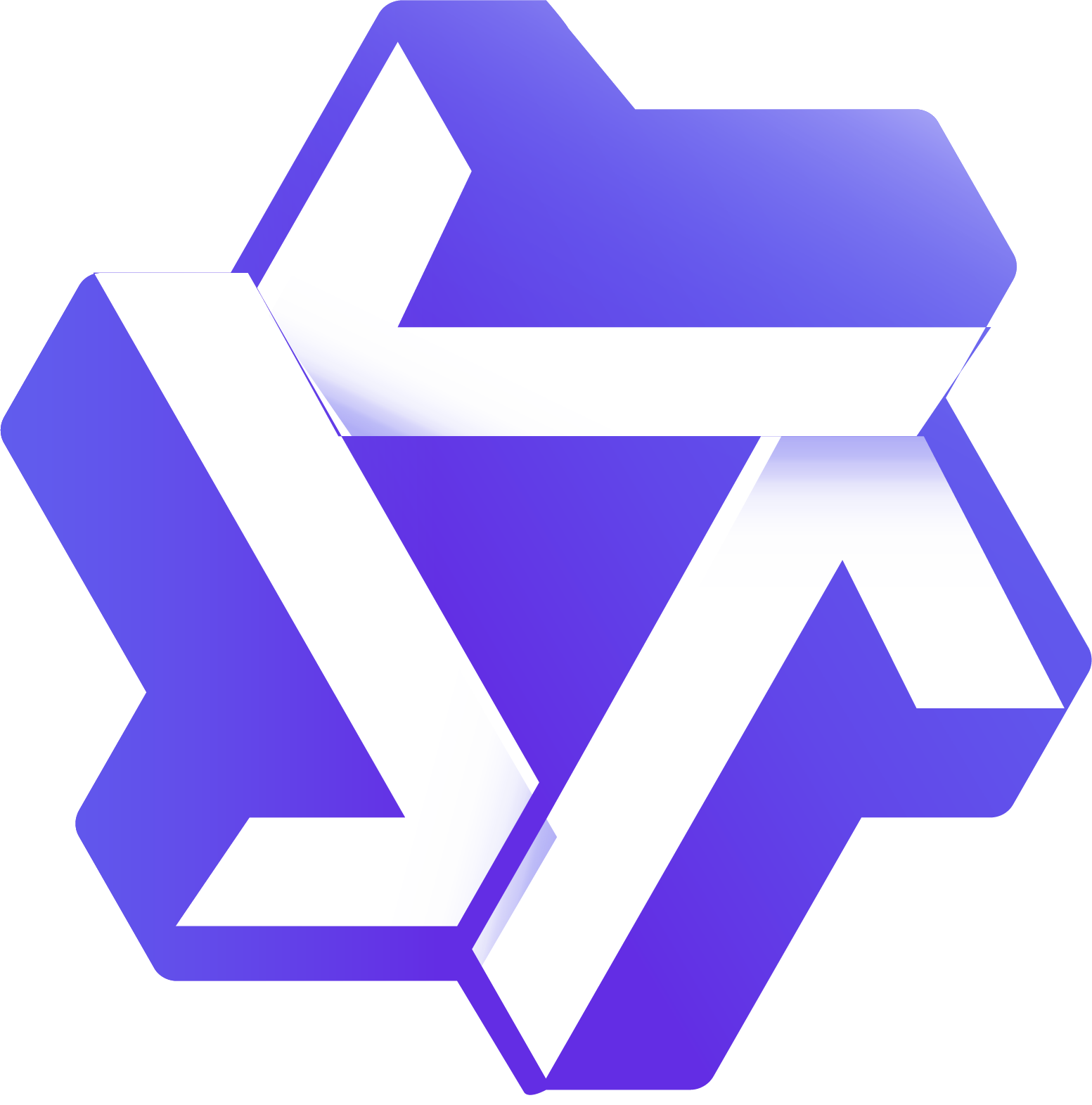}}}
\newcommand{\iconllama}{\raisebox{-0.15em}{\includegraphics[height=1.00em]{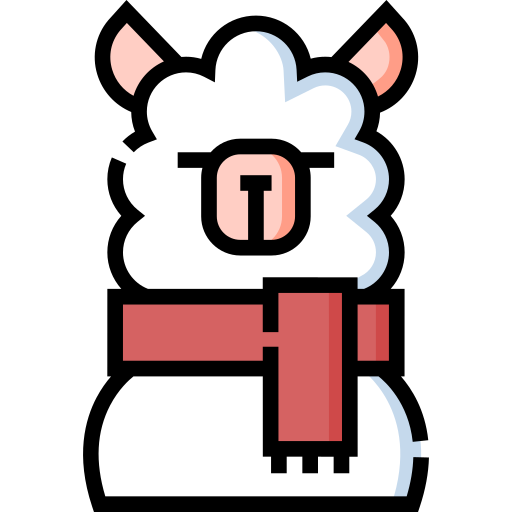}}}

\lstdefinestyle{plain}{
    basicstyle=\fontsize{7}{9.5}\ttfamily,
    keywordstyle=\color{blue},
    commentstyle=\color{gray},
    stringstyle=\color{green},
    showstringspaces=false,
    breaklines=true,
    breakatwhitespace=false,
    breakindent=0pt,
    escapeinside={(*@}{@*)}
}

\usepackage{booktabs}
\usepackage{makecell}
\usepackage{graphicx}
\usepackage[table]{xcolor}

\definecolor{posgreen}{HTML}{5AA469}
\definecolor{negred}{HTML}{D36B6B}
\newcommand{\up}[2]{\cellcolor{posgreen!#1}#2}
\newcommand{\down}[2]{\cellcolor{negred!#1}#2}

\begin{document}
\maketitle
\renewcommand{\thefootnote}{\fnsymbol{footnote}}
\footnotetext[2]{Corresponding author.}
\renewcommand{\thefootnote}{\arabic{footnote}}
\footnotetext[1]{Data: 
\href{https://github.com/yiwenJG/VIBE-Bench}{\url{https://github.com/yiwenJG/VIBE-Bench}}}

\begin{abstract}
Personalized Large Language Models (PLLMs) aim to tailor responses to individual users, where a central challenge is preference reasoning: inferring query-relevant preferences from user-related history. Existing benchmarks, however, largely assume that such preference can be retrieved from semantically related history. We study an underexplored but practically important regime, profile-preference conceptual misalignment (PRCM), where observable profile cues and query-specific preferences lie in different concept spaces, making semantic retrieval inconsistent for personalization. We introduce \textsc{Vibe-Bench}\footnotemark{}, a benchmark with two psychology-grounded tasks, 3,504 personas and 12,239 dialogues, including a manually verified gold test set, and requires cross-concept preference reasoning beyond surface semantic overlap. Experiments with several personalization methods show that current PLLMs largely rely on shallow semantic correlations and fail to acquire robust cross-concept mappings. These findings establish PRCM as a distinct failure regime in PLLMs and position \textsc{Vibe-Bench} as a focused testbed for advancing preference reasoning beyond semantic matching.

\end{abstract}

\section{Introduction}

Large Language Models (LLMs) have enabled general-purpose systems across diverse NLP tasks \cite{touvron2023llamaopenefficientfoundation, zhang-etal-2025-badwindtunnel}, yet they are predominantly trained under a \textit{one-size-fits-all} paradigm that limits adaptation to individual users \cite{wu-etal-2021-personalized,zhao2025surveylargelanguagemodels}. To address this limitation, Personalized Large Language Models (PLLMs) leverage user-specific data, e.g., interaction histories, to provide \textit{one-size-fits-one} responses \cite{liu2025surveypersonalizedlargelanguage}. Recently, numerous benchmarks have been proposed to evaluate how well PLLMs align with user preferences (\citealp{jiang2025knowmerespondme}; \citealp{ong-etal-2025-towards}; \citealp{au2025personalizedgraphbasedretrievallarge}).

A central challenge in PLLMs is \textbf{preference reasoning}: inferring, from a user’s historical data, which preferences should govern the response to the current query \cite{zhang2025personalization}. Existing personalization methods largely approach this problem through semantic matching between the query and user history. Consequently, mainstream paradigms, including retrieval-augmented prompting \cite{zhang-etal-2024-llm-based, kim2026rpm}, profile-augmented prompting \cite{richardson2023integratingsummarizationretrievalenhanced, qiu-etal-2025-measuring}, and personalized fine-tuning \cite{tan-etal-2024-democratizing}, typically first locate semantically related historical evidence, and then personalize the response accordingly. Even recent work on implicit preference reasoning \cite{wu-etal-2025-aligning} largely remains within this paradigm. For example, IMPLEXCONV \cite{li-etal-2025-toward} defines harder implicit reasoning cases by increasing semantic distance through predefined thresholds, but still evaluates preference reasoning within weak semantic alignment rather than extending beyond it.

\begin{figure*}[t]
  \centering
  \includegraphics[width=\textwidth]{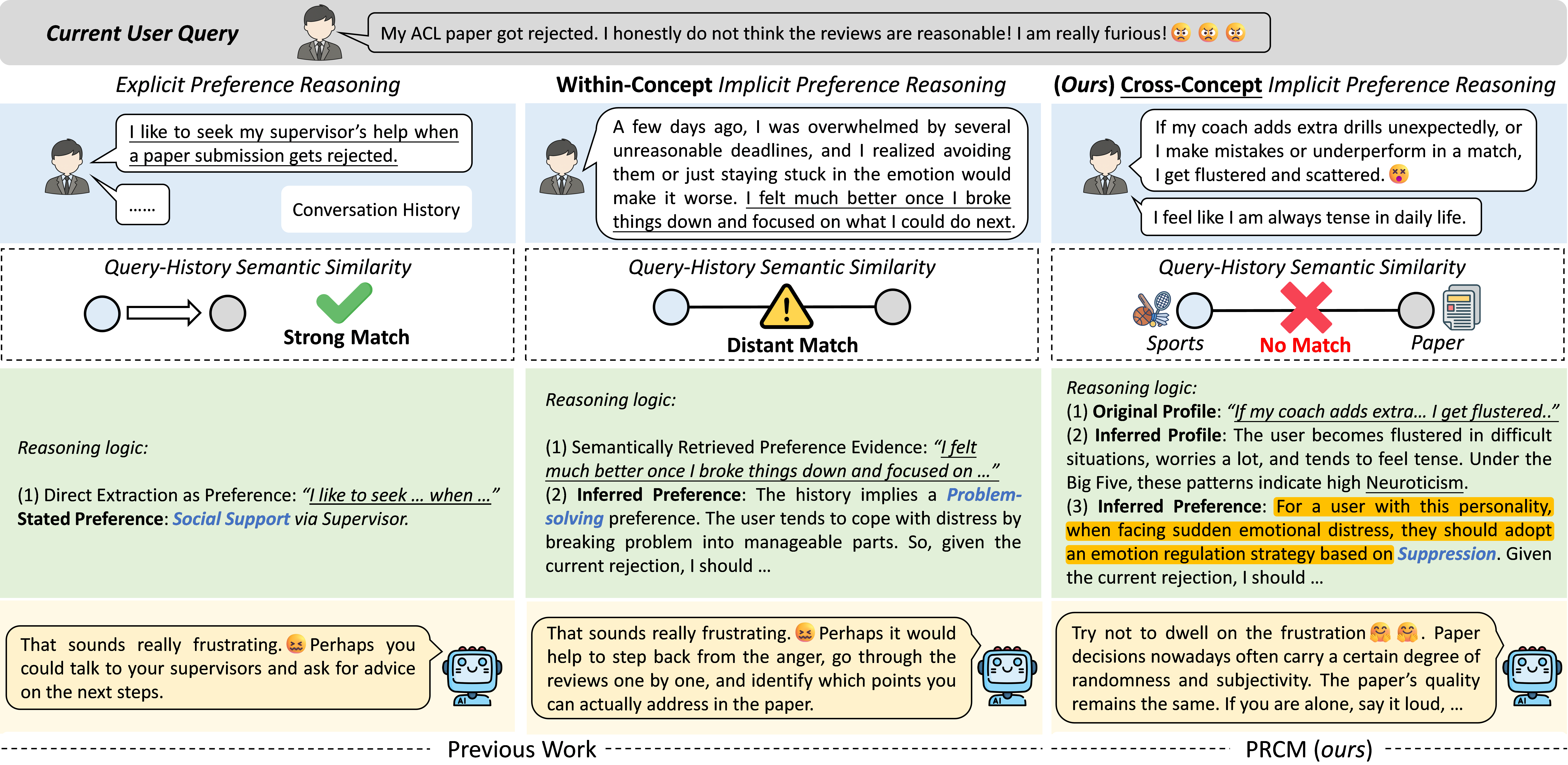}
  \caption{\textbf{Illustration of three preference-reasoning regimes in PLLMs}. In \emph{cross-concept implicit preference reasoning} (our focus, PRCM), observable profile cues and target preferences lie in different concept spaces, requiring inference through cross-concept mappings rather than the direct semantic matching assumed in previous work.}
  \label{fig:intro}
\end{figure*}

The current paradigm assumes \textbf{preference retrievability}: the preferences relevant to the current query can be recovered from semantically related history. In realistic settings, however, this assumption often fails. Users may not fully understand their own preferences \cite{annurev:/content/journals/10.1146/annurev.psych.55.090902.141954}, and they rarely mention them in a direct or query-aligned way \cite{li-etal-2025-toward}. The cold-start problem in personalization \cite{au2025personalizedgraphbasedretrievallarge, li2025personalizedreasoningjustintimepersonalization} is one concrete manifestation of this issue. More generally, personalization often begins from indirect, weak, or semantically mismatched evidence. In such cases, better retrieval is not enough: PLLMs must infer from clues that do not directly resemble the target query or response style.

We study this setting, where the historical cues available in a user profile and the target preference required by the current query reside in different concept spaces. We call this phenomenon \underline{P}rofile-p\underline{R}eference \underline{C}onceptual \underline{M}isalignment (PRCM). Figure~\ref{fig:intro} illustrates the distinction. In the example, effective emotion-regulation guidance should be personalized according to the user’s personality traits, which must be inferred from interaction history. Yet the cues that reveal personality, such as sports-related topics, are not semantically aligned with emotion-regulation strategies. Their connection instead arises from an underlying cross-concept regularity supported by established psychological theory \cite{baranczuk2019five}. PRCM therefore requires models to do more than retrieve semantically related evidence. They must bridge profile cues and target preferences across concept spaces. Although pervasive in real-world personalization, PRCM remains largely unevaluated.

\begin{figure}[t]
  \centering
  \includegraphics[width=0.8\columnwidth]{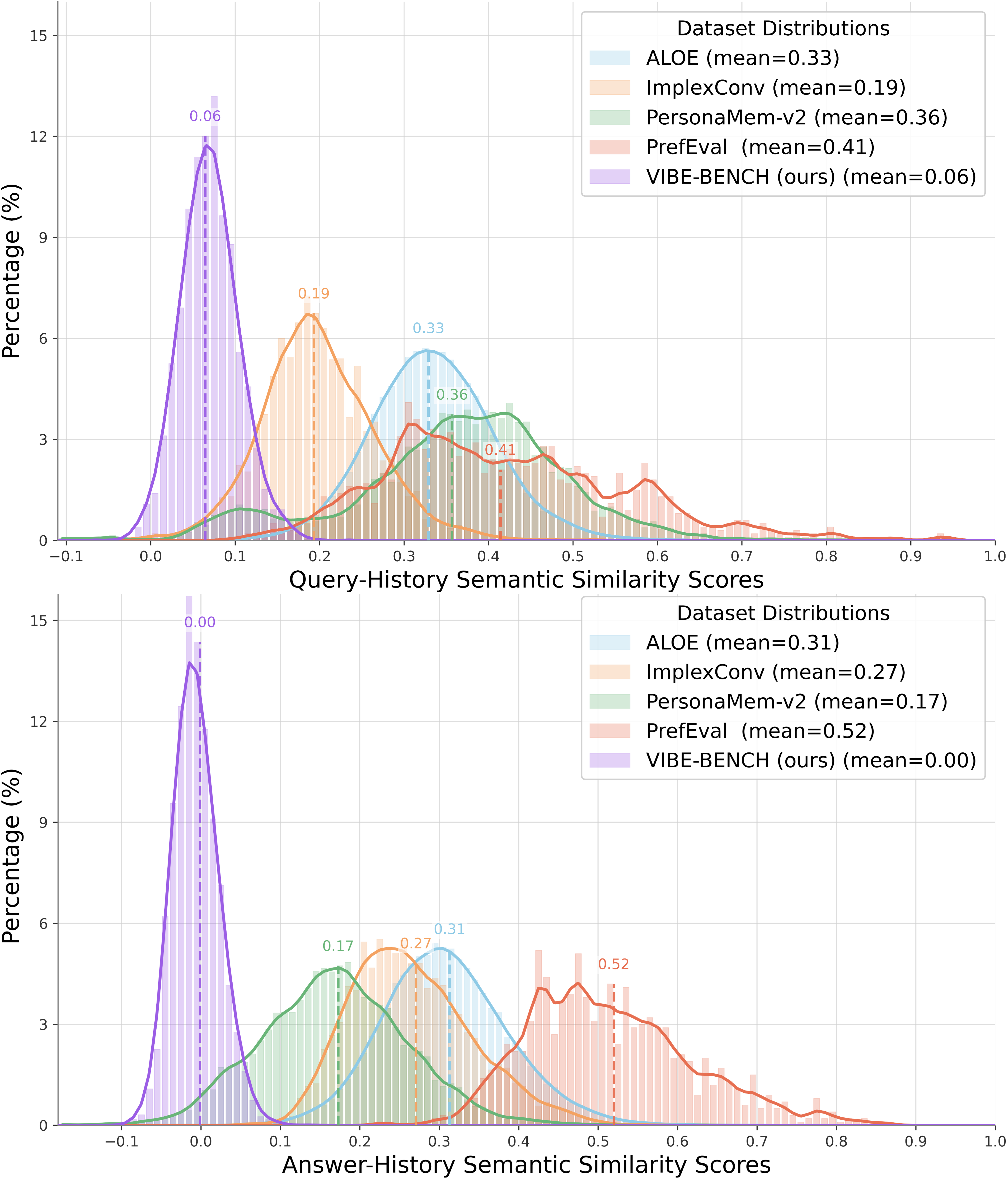}
  \caption{Semantic similarity distributions of \textsc{Vibe-Bench} and prior implicit preference reasoning benchmarks. The x-axis is semantic similarity, and the y-axis is the percentage of samples at each similarity level.}
  \label{fig:sim}
\end{figure}

To address this gap, we introduce \textsc{Vibe-Bench} (\textit{\underline{V}ocational \underline{I}nterests and \underline{B}ig-Five-based \underline{E}motion-Regulation \underline{Bench}mark}), a benchmark designed to isolate preference reasoning under PRCM, comprising 3,504 personas and 12,239 dialogues. We instantiate PRCM through two psychology-grounded tasks that naturally require cross-concept inference: (i) \textbf{Emotion-Regulation Generation}, in which a PLLM infers personality cues related to the Big Five \citep{mccrae1992introduction}, identifies an appropriate emotion-regulation strategy, and generates a supportive response for a distress scenario; (ii) \textbf{Vocational Interest Classification}, in which a PLLM judges whether a user’s occupation is consistent with their inferred interest structure under Holland’s RIASEC theory \citep{holland1997making}. We choose these tasks because they provide controlled, well-grounded testbeds where profile cues and target preferences are known to reside in different concept spaces. To validate the distinctiveness of \textsc{Vibe-Bench}, we compare it quantitatively with representative prior benchmarks for implicit preference reasoning, using semantic similarity between the user history and both the query and the response. Figure~\ref{fig:sim} shows that \textsc{Vibe-Bench} is near zero on both measures, confirming a substantially higher level of implicitness (see Appendix~\ref{appendix:semantic_similarity}). Based on \textsc{Vibe-Bench}, we adapt PLLMs with several personalization methods. We find that standard personalization-oriented fine-tuning \cite{tan-etal-2024-democratizing} primarily reinforces shallow surface correlations, rather than learning the cross-concept mappings required for robust personalization. This makes the automatic discovery of such mappings a central open challenge for PLLMs. Our main contributions are as follows:
\begin{itemize}[leftmargin=*,nosep]
    \item We identify PRCM as a distinct and underexplored preference-reasoning regime in PLLMs, and propose a novel taxonomy that characterizes it as cross-concept implicit preference reasoning beyond semantic retrieval.
    \item We build \textsc{VIBE-Bench}, a benchmark that isolates PRCM through two psychology-grounded tasks with 3,504 personas and 12,239 dialogues.
    \item We show that current personalization methods favor semantic correlations over cross-concept profile-preference mappings, revealing conceptual mapping as a key bottleneck under PRCM.
\end{itemize}

\section{Related Work}

\noindent \textbf{Benchmarks for Personalized LLMs.} Numerous benchmarks (\citealp{wu-etal-2021-personalized}; \citealp{salemi-zamani-2025-lamp}; \citealp{qiu-etal-2025-measuring}; \citealp{zhao-etal-2025-personalens}) have been proposed to evaluate PLLMs' ability to align with user preferences. 
LaMP \citep{salemi-etal-2024-lamp} assesses preference induction through classification and generation over users' historical behaviors, while PersonaBench \citep{li2025personalizedconversationalbenchmarksimulating} and PersonalLLM \citep{DBLP:conf/iclr/ZolloSYLN25} evaluate persona- and preference-consistent responses via multi-turn dialogue and open-ended prompts with reward modeling. Subsequent benchmarks emphasize long-context and memory-based personalization (\citealp{kumar2024longlampbenchmarkpersonalizedlongform}; \citealp{du-etal-2024-perltqa}; \citealp{maharana-etal-2024-evaluating}; \citealp{lee2025realtalk21dayrealworlddataset}; \citealp{tan-etal-2025-membench}; \citealp{wang2026memguardpersistingverifiersignals}; \citealp{zhang2026memmarkstateevolutionattributionwatermarking}), as well as cold-start and sparse histories (\citealp{au2025personalizedgraphbasedretrievallarge}; \citealp{li2025personalizedreasoningjustintimepersonalization}), continual personalization (\citealp{jiang2025knowmerespondme}; \citealp{ong-etal-2025-towards}), and multimodal or task-oriented settings, e.g., conversational assistants \citep{mok-etal-2025-exploring}, web agents \citep{cai2025large}, and recommender systems \cite{10.1145/3701716.3717531}. Despite their breadth, these benchmarks assume that user preferences can be inferred from historical evidence, leaving the evaluation of conceptual misalignment unexplored.

\noindent \textbf{Implicit Preference Reasoning.}
Recent benchmarks move beyond explicit profile conditioning \citep{DBLP:conf/iclr/WuWYZCY25} to evaluate implicit preference inference from scattered cues across multi-turn interactions. ALOE \citep{wu-etal-2025-aligning} prompts PLLMs to progressively infer preferences from signals such as linguistic style and topic choice, but still largely treats preferences as reducible to profile cues. IMPLEXCONV \citep{li-etal-2025-toward} distributes preference-relevant evidence across syntactically subtle and semantically distant contexts, requiring multi-stage summarization and retrieval, while PrefEval \citep{DBLP:conf/iclr/Zhao00HL25} and PersonaMem-v2 \citep{jiang2025personamemv2personalizedintelligencelearning} evaluate models' ability to infer, retain, and apply implicit preferences in long-context and multi-user settings. However, across these benchmarks, preference targets and supporting evidence typically reside in a shared semantic space, allowing similarity-based retrieval to bridge them directly and leaving cross-concept preference reasoning largely unexplored. PRISM \citep{DBLP:conf/nips/KirkWRBMGCBW0VH24} takes an initial step toward modeling conceptual mismatch by jointly analyzing group-level preferences and broad individual attributes, but it is not designed to systematically evaluate PRCM.

\section{Task Formulation}
\label{sec:3}

Let $\mathcal{U}=\{u_i\}_{i=1}^{N}$ be a set of users, and let $\mathcal{P}$, $\mathcal{Q}$, $\mathcal{Y}$, and $\mathcal{R}$ denote the spaces of user profiles, queries, responses, and (query-conditioned) preferences, respectively.
For each user $u_i$, let $p_i \in \mathcal{P}$ be their profile. For their $j$-th query $q^{(i)}_j \in \mathcal{Q}$, denote the corresponding response by $y^{(i)}_j \in \mathcal{Y}$.

\noindent \textbf{Profile.} The profile $p_i$ encodes user-specific prior information available before the current query. It is query-independent, shared across all queries from $u_i$ as a reusable long-term representation.

\noindent \textbf{Preference.} For each query $q^{(i)}_j$, let $r^{(i)}_j \in \mathcal{R}$ denote the corresponding (target) preference, i.e., the query-specific constraints and objectives that should guide personalized generation or decision-making. Since it is tied to the current query, $r^{(i)}_j$ may vary across queries for the same user.

\noindent \textbf{Task Objective.} A non-personalized LLM implements a mapping $f_{\text{LLM}}:\mathcal{Q}\rightarrow\mathcal{Y}$. In contrast, a PLLM conditions on both the query and the profile, implementing $f_{\text{PLLM}}:\mathcal{Q}\times\mathcal{P}\rightarrow\mathcal{Y}$:
\[
y^{(i)}_j = f_{\text{PLLM}}\bigl(q^{(i)}_j,\, p_i\bigr)
\]
where $y^{(i)}_j$ is expected to align with user $u_i$'s preferences given their profile $p_i$.

\section{Proposed Taxonomy}

We provide definitions of concepts and concept spaces in the context of this work in Appendix~\ref{appendix:concept}.

\paragraph{Conceptual Misalignment.} PLLMs operate under \emph{Profile-Preference Conceptual Misalignment} (PRCM) when, for a query $q^{(i)}_j$, the evidence available in $p_i$ is not expressed in the concept space required to determine $r^{(i)}_j$. In this paradigm, retrieval over $p_i$ based on semantic similarity to $q^{(i)}_j$ cannot directly surface concept-level evidence for $r^{(i)}_j$. To obtain an initial preference $r_0$, the model must first map profile-level concepts to query-relevant preference concepts, denoted as
\[
p_\ast \xrightarrow{\phi} r_0,
\]
where $p_\ast$ is a (possibly query-conditioned) preprocessing of $p_i$, and $\phi$ may leverage external knowledge, population-level statistics, or aligned ontologies. The resulting $r_0$ can then be further refined to $r_\ast$ as in the aligned setting. We provide the definition of conceptual alignment in Appendix~\ref{appendix:ca}.

We propose a technical taxonomy of preference reasoning with three paradigms (Figure~\ref{fig:intro}):

$\bullet$ \textbf{Explicit Preference Extraction}. The target preference is explicitly encoded in the profile and can be obtained via direct evidence retrieval and extraction; the initial preference $r_0$ already suffices.

$\bullet$ \textbf{Within-Concept Implicit Preference Reasoning}. The target preference is not explicitly stated but is inferable from conceptually co-level evidence in the profile. The model must identify semantically distant yet conceptually related cues to form an initial estimate $r_0$ and refine it into a query-appropriate preference $r_\ast$ via aggregation, induction, or generalization:
\[
r_0 \rightarrow r_\ast
\]

$\bullet$ \textbf{Cross-Concept Implicit Preference Reasoning}. Under PRCM, query-relevant preferences cannot be retrieved by semantic matching alone, because the profile evidence and the target preference reside in different concept spaces. The model must therefore apply a cross-space mapping $\phi$ that links profile-level concepts to preference-level concepts:
\[
p_0 \rightarrow p_\ast \xrightarrow{\phi} r_0 \rightarrow r_\ast
\]
where the query-conditioned preprocessing ($p_0 \rightarrow p_\ast$) and refinement ($r_0 \rightarrow r_\ast$) steps are optional and instantiated only when needed.

\section{\textsc{Vibe-Bench} Construction}

\subsection{Cross-Concept Design Principles}

The \textsc{Vibe-Bench} instantiates two PRCM cases grounded in established psychological theories: First, the Big Five personality model \citep{mccrae1992introduction} characterizes stable behavioral, cognitive, and affective patterns along five dimensions. Emotion regulation comprises processes by which individuals alter how they experience and express emotion through cognitive and behavioral strategies \cite{gross1998emerging, gratz2004multidimensional, pena2015integrating}. Substantial empirical evidence shows that Big Five traits are systematically associated with different emotion regulation strategies, with heterogeneous directions and magnitudes of correlation across traits and strategies \cite{baranczuk2019five}. Second, Holland's RIASEC framework \citep{holland1997making} is a theory of vocational interest-occupation fit that characterizes which types of jobs and work environments best match an individual's interests. It groups vocational interests into six types. Details in Appendix~\ref{sec:scyt}.

We adopt these two theories because they capture natural profile-preference misalignment, specifically personality-emotion regulation (Table~\ref{tab:personality-strategy}) and occupation-interest (Table~\ref{tab:job-riasec}). Moreover, Big Five and RIASEC traits are rarely stated explicitly in everyday dialogue, yielding an implicit preference-reasoning setting that requires inductive profile inference from contextual cues.

\subsection{Metadata Collection}

\noindent \textbf{Big Five Personality.} We construct Big Five personality metadata from two sources: behavior-level descriptions derived from public personality inventories and language-pattern features from \textsc{BIG5-CHAT}. This results in 696 high/low trait descriptions, complemented by psycholinguistic data that captures personality-related language-use patterns.

\noindent \textbf{Occupational Information Network.}
We curate occupational metadata from O$^{*}$NET\footnote{\url{https://www.onetonline.org/}}\label{fn:onet}, covering 876 jobs across 22 major groups. Each entry includes structured information, e.g., job titles, work duties, and corresponding RIASEC interest categories.

\noindent \textbf{Holland's RIASEC Framework.}
We use the \textit{O$^{*}$NET Interest Profiler (Long Form)}\footnotemark[\value{footnote}] to construct RIASEC interest metadata, covering 180 self-descriptive activity-preference items associated with the six vocational interest domains.

\noindent \textbf{Emotional Distress Event.} We sample 1,142 and 2,362 instances, respectively, from two emotional-support dialogue datasets, ESConv \cite{liu-etal-2021-towards} and ExTES \cite{zheng2023buildingemotionalsupportchatbots}, using stratified sampling over event categories. More details on metadata collection are provided in Appendix~\ref{appendix:meta}.

\subsection{Dataset Generation}

\begin{table}[t]
\centering
\resizebox{0.90\columnwidth}{!}{%
\begin{tabular}{ccccc}
\Xhline{1pt}
\multicolumn{2}{c}{\textbf{Data Statistics}}                                                                                                                                                & \textbf{Train} & \textbf{Valid} & \textbf{Test} \\ \Xhline{1pt}
\multirow{3}{*}{\#Dialogue}  & \iconbigfiveup                                                                                       & 2,532   & 844     & 128    \\
                           & \iconbigfivedown                                                                                     & 1,886   & 630     & 90     \\
                           & \iconriasec                                                                                          & 4,437   & 1,468   & 224    \\ \hline
\multirow{3}{*}{\#Utterance} & \iconbigfiveup                                                                                       & 26,364  & 8,688   & 1,312  \\
                           & \iconbigfivedown                                                                                     & 20,640  & 6,906   & 982    \\
                           & \iconriasec                                                                                          & 47,270  & 15,698  & 2,392  \\ \hline
\multirow{4}{*}{\begin{tabular}[c]{@{}c@{}}\#Session\\ Pattern\end{tabular}}   & \iconbigfiveup~\iconbigfivedown~\iconriasec                            & 485     & 173     & 24     \\
                           & \iconbigfiveup~\iconriasec                                                            & 142     & 47      & 8      \\
                           & \iconbigfiveup~\iconbigfivedown~\iconriasec~\iconriasec & 1,401   & 457     & 66     \\
                           & \iconbigfiveup~\iconriasec~\iconriasec                                 & 504     & 167     & 30     \\ \hline
\multicolumn{2}{c}{\#Persona}                                                                                                                      & 2,532   & 844     & 128    \\ \Xhline{1pt}
\end{tabular}%
}
\caption{\textsc{Vibe-Bench} Statistics. \iconbigfiveup, \iconbigfivedown, \iconriasec~denote sessions encoding high Big Five traits, low Big Five traits, and RIASEC-related evidence, respectively. \textit{Session Pattern} reports session-type counts (one icon per occurrence); session order is shuffled in the dataset.}
\label{tab:sta}
\end{table}

For each occupation, we construct four samples. In each sample, one Big Five dimension is set to a high level, paired with one or two dimensions set to low levels. Two samples match the occupation's RIASEC interest type, and two are intentionally mismatched, yielding a balanced split. In total, \textsc{Vibe-Bench} contains 3,504 samples.

\noindent \textbf{Step 1. Persona Card Generation.} Each sample is paired with a persona card that remains fixed throughout data synthesis. The card specifies name, age, gender, job title, education level, Big Five traits, self-described job content, and interest activities aligned with the target RIASEC type.

\texttt{GPT-5-mini} instantiates these fields under constraints derived from the collected metadata. To increase diversity and refine the self-descriptions, we sample three incumbent-reported job titles and four work duties from each occupation's O\textsuperscript{*}NET entry, along with four self-descriptive items for the target RIASEC dimension from the \textit{Interest Profiler}. Conditioned on these cues, \texttt{GPT-5-mini} generates two self-described job responsibility statements and two self-described interest activities consistent with the target RIASEC type.

\noindent \textbf{Step 2. Work-Interest Event Generation.} Given the self-described job responsibilities and interest activities, we generate contextualized event instances with explicit settings and action sequences. To instantiate the target Big Five trait in work events, we condition \texttt{GPT-5} on the persona card, the trait definition, and four behavioral cues sampled from the personality inventory. \texttt{GPT-5} is instructed to weave these cues into the event progression so that each work event reflects either the high or low level of the target trait dimension. This step yields two work events (high vs.\ low) and two interest events aligned with the same target interest type.

\noindent \textbf{Step 3. Historical Dialogue Generation.} We convert the generated events into multi-turn dialogues between a user and a chatbot, using the resulting dialogue histories as user profiles. We prompt \texttt{GPT-5-mini} to let the chatbot elicit event details through guided questions, such that the user reveals context and key details gradually over multiple turns rather than in a single narrative. We also instruct \texttt{GPT-5-mini} to surface cues indicative of the user's personality traits and interest preferences. To simulate user-specific speaking styles (e.g., lexical choices, tone), we include psycholinguistic cues in the prompt, encouraging consistency with the specified high or low level of the target trait.

Under this pipeline, each profile contains up to four dialogue sessions. To diversify profile length, we enforce sampling constraints: each profile includes at least one session reflecting the target personality dimension (high level) and one session reflecting the target interest type, while the remaining sessions are randomly sampled from candidate dialogues. As a result, profiles contain 2, 3, and 4 sessions in 5.62\%, 39.47\%, and 54.91\% of cases, respectively (see Table~\ref{tab:sta} for detailed statistics).

\noindent \textbf{Step 4. Emotional Distress Event Q\&A.} Conditioned on the persona card and the metadata-specified distress event, \texttt{GPT-5-mini} generates a user-specific distress scenario as the query (time, involved parties, developments, and post-event emotional reactions). We restrict candidate emotion-regulation strategies via a predefined personality-strategy mapping, and have \texttt{GPT-5-mini} select a strategy that is both event-appropriate and trait-consistent. Given the event, the selected strategy definition, and application guidance, \texttt{GPT-5-mini} produces the corresponding emotion-regulation response. We thus obtain one Q\&A pair per sample with preference signals that are controllable and interpretable through the explicit mapping constraint.

\subsection{Task Definitions}

The \textsc{Vibe-Bench} includes two tasks with shared user profiles but distinct queries. Turns relevant to one task serve as distractors for the other, increasing the challenge of preference reasoning.

\noindent \textbf{Task 1. Emotion-Regulation Generation.} Given a user's dialogue history, the model extracts behavioral and psycholinguistic cues ($p_0$) and induces the most salient personality dimension ($p_0 \rightarrow p_\ast$). It then uses a personality-strategy mapping to derive a candidate strategy set ($p_\ast \xrightarrow{\phi} r_0$). Conditioned on a current emotional distress event, the model selects the best-matched strategy and generates an event-tailored, guided emotion-regulation response that instantiates the selected strategy ($r_0 \rightarrow r_\ast$).

\noindent \textbf{Task 2. Vocational Interest Classification.} Given a user's dialogue history, the model extracts self-described job responsibilities ($p_0$) and induces a specific job title ($p_0 \rightarrow p_\ast$). It then derives the corresponding RIASEC interest type via a job-RIASEC mapping ($p_\ast \xrightarrow{\phi} r_0$). Finally, it compares this type with the RIASEC type of the user's actual leisure activities mentioned in the dialogue history ($r_0 \rightarrow r_\ast$) and predicts whether they match (matched vs. mismatched).

\subsection{Dataset Splits and Statistics}

We reserve 32 occupations for testing, disjoint from those used for training and validation. For the remaining occupations, we apply an occupation-stratified split by sampling one instance per occupation for validation and assigning the rest to training. Table~\ref{tab:sta} reports the final distribution. The test set comprises 128 manually verified and corrected samples, which serve as the gold standard for evaluation. Overall, \textsc{Vibe-Bench} contains 12,239 dialogues, comprising 130K utterances. On average, each dialogue contains 10.6 utterances, and each user profile spans 2-4 dialogue sessions.



\begin{figure}[t]
  \centering
  \includegraphics[width=1.0\columnwidth]{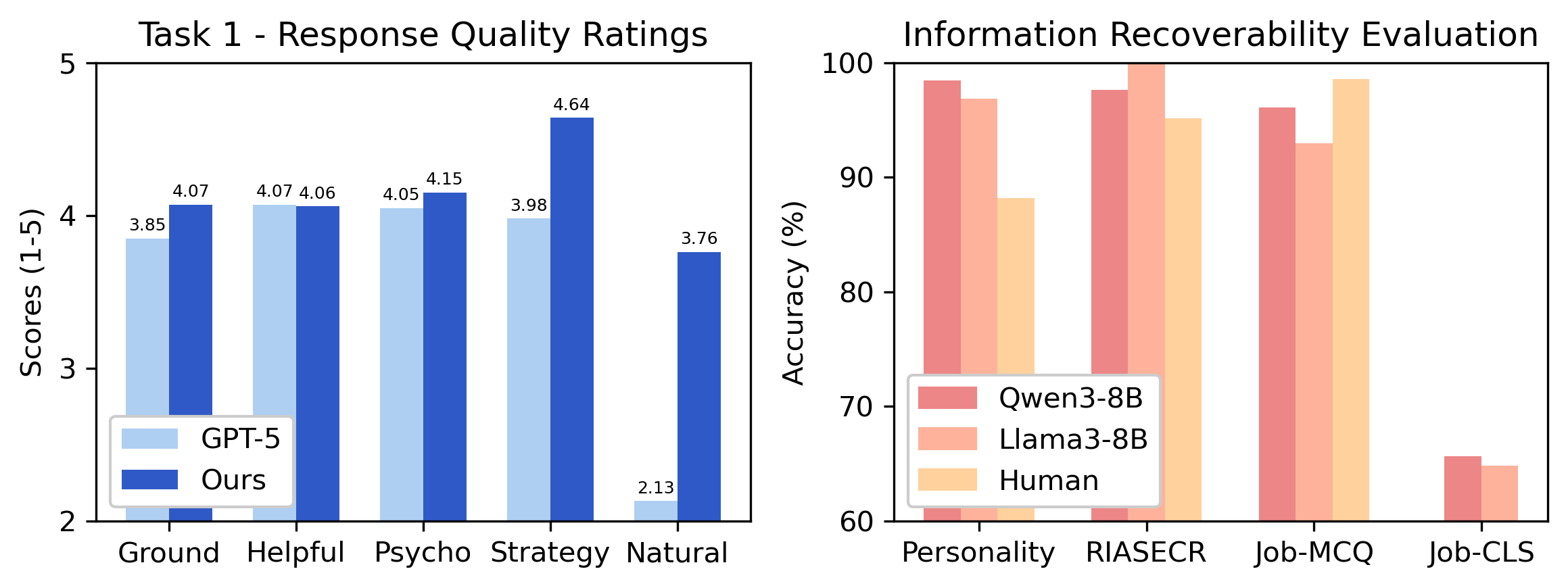}
  \caption{Dataset Quality Assessment. \textit{Left}: Five-dimensional evaluation of Task 1 generation quality. \textit{Right}: Recovery of profile information from dialogue histories. Job-MCQ uses four sampled distractor occupations, while Job-CLS is candidate-free.}
  \label{fig:vibe-quality}
\end{figure}

\begin{table*}[t]
\centering
\small
\setlength{\tabcolsep}{3pt}
\renewcommand{\arraystretch}{1.10}

\resizebox{0.90\textwidth}{!}{%
\begin{tabular}{c|cccc|cccc|ccccc}
\toprule
\multicolumn{1}{c|}{}
& \multicolumn{8}{c|}{\textbf{Task 1}}
& \multicolumn{5}{c}{\textbf{Task 2}} \\
\cmidrule(lr){2-9} \cmidrule(lr){10-14}

\textbf{Method}
& \multicolumn{4}{c|}{\textbf{Personalized Strategy}}
& \multicolumn{4}{c|}{\textbf{Personalized Response}}
& \multicolumn{5}{c}{\textbf{Personalized Classification}} \\
\cmidrule(lr){2-5} \cmidrule(lr){6-9} \cmidrule(lr){10-14}

& ACC
& B-ACC
& M-F1
& W-F1
& BLEU
& R-L
& MET
& BS-F1
& ACC
& PRE
& REC
& F1
& MCC \\
\midrule

Base-LLM
& 21.48
& 19.81
& 14.15
& 16.91
& 3.10
& 17.46
& 24.73
& 8.84
& 50.00
& 50.00
& 50.00
& 50.00
& 0.00 \\

FHP-LLM
& \down{41}{18.36}
& \down{41}{17.20}
& \down{45}{10.36}
& \down{43}{12.48}
& \down{10}{3.03}
& \down{12}{17.04}
& \down{38}{21.89}
& \down{13}{8.41}
& \up{21}{60.81}
& \up{21}{61.74}
& \up{31}{67.97}
& \up{23}{62.73}
& \up{22}{0.23} \\

PAP-LLM
& \up{37}{24.22}
& \up{34}{21.85}
& \up{42}{17.61}
& \up{37}{20.49}
& \up{11}{3.35}
& \up{11}{17.63}
& \up{18}{25.53}
& \down{11}{8.68}
& \up{23}{62.24}
& \up{19}{59.88}
& \up{41}{76.30}
& \up{27}{66.50}
& \up{24}{0.27} \\

RAP-BM25
& 21.48
& \down{22}{18.82}
& \up{15}{14.66}
& \up{17}{17.78}
& \up{10}{3.15}
& \up{10}{17.51}
& \down{18}{23.96}
& \up{12}{9.10}
& \up{16}{55.47}
& \up{15}{55.50}
& \up{20}{58.85}
& \up{17}{56.39}
& \up{16}{0.11} \\

RAP-BERT
& \up{22}{22.66}
& \up{14}{20.18}
& \up{21}{15.39}
& \up{21}{18.38}
& \up{10}{3.11}
& \up{10}{17.54}
& \down{16}{24.10}
& \up{11}{9.08}
& \up{16}{55.73}
& \up{14}{54.23}
& \up{36}{72.40}
& \up{22}{61.59}
& \up{17}{0.13} \\

P-SFT
& \down{45}{17.97}
& \down{45}{16.88}
& \down{44}{10.46}
& \down{45}{12.23}
& \up{45}{9.11}
& \up{45}{26.65}
& \up{45}{28.25}
& \up{45}{14.84}
& \up{45}{83.98}
& \up{45}{87.28}
& \up{45}{79.69}
& \up{45}{83.15}
& \up{45}{0.68} \\

\bottomrule
\end{tabular}%
}

\caption{Main Results on \textsc{Vibe-Bench}. Base-LLM is used as the reference row. Green indicates improvement, while red indicates degradation. Darker colors denote larger differences. See Table~\ref{tab:task1_results} and Table~\ref{tab:task2_results} for full results.}
\label{tab:main_results}
\end{table*}

\section{\textsc{Vibe-Bench} Quality Assessment}

\subsection{Information Recoverability}
We assess whether the multi-session profiles contain sufficient cues to recover three target attributes: (i) the dominant Big Five trait, (ii) the RIASEC interest type, and (iii) the user's occupation.

\noindent \textbf{Model-Based.} We fine-tune \iconqwen~\texttt{Qwen3-8B} \cite{yang2025qwen3technicalreport} and \iconllama~\texttt{Llama3-8B} \cite{grattafiori2024llama3herdmodels} to test if the profiles provide discriminative evidence. In Figure~\ref{fig:vibe-quality}, both models attain high accuracy on \textit{Personality} and \textit{RIASEC} (>96\%). \textit{Job-CLS} drops to 65\%, likely because all test occupations are unseen and require cross-occupation generalization. These results indicate most persona attributes are recoverable from profile text, while occupation prediction is harder under distribution shift.

\noindent \textbf{Human Annotation.} Three annotators answered four multiple-choice questions for each test instance. They first annotated a shared set of eight instances to estimate inter-annotator agreement (Fleiss' $\kappa=0.84$), and then independently annotated 40 instances each. Figure~\ref{fig:vibe-quality} reports accuracy against the gold persona cards. Human accuracy on personality inference is lower than that of LLMs, likely because the protocol emphasizes behavioral evidence, whereas LLMs can also exploit fine-grained linguistic cues. A fourth annotator adjudicated disagreements, after which we revised six ambiguous segments to remove confounders and better align them with the intended labels.

\subsection{Response Quality Assessment}

We evaluate Task 1 responses along five dimensions: \textit{Helpfulness}, \textit{Psychological Appropriateness}, \textit{Conversational Naturalness}, \textit{Event Grounding}, and \textit{Strategy Realization} in Figure~\ref{fig:vibe-quality} (see Appendix~\ref{appendix:rubric} for rubric). As a baseline, we use the same \texttt{GPT-5-mini} to directly generate responses without the expert-designed prompt template.

After confirming strong agreement between expert  and Claude, we use \texttt{Claude-Haiku-4.5} to score 256 samples. The expert-designed prompt template consistently improves response quality. Compared with vanilla \texttt{GPT-5-mini}, our method yields the largest gains in \textit{Naturalness} and \textit{Strategy Realization}, and also improves \textit{Event Grounding} and \textit{Psychological Appropriateness}. The average score across all five dimensions exceeds 4, indicating that our generated responses closely resemble realistic emotional support scenarios.

\section{Experiments}

\subsection{Baselines}
We evaluate four groups of methods on \textsc{Vibe-Bench}. (1) Non-Personalized LLM (Base-LLM) uses only the current query, without access to user history, serving as the baseline. (2) Full-History Prompting (FHP-LLM) concatenates the entire user history into the prompt to test whether long-context history improves personalization. (3) Non-Parametric Personalization augments the prompt with external user information. We consider the following variants: Profile-Augmented Prompting (PAP-LLM) \cite{richardson2023integratingsummarizationretrievalenhanced, qiu-etal-2025-measuring}, which adds a long-term user profile summarized from past dialogues, and Retrieval-Augmented Prompting \cite{zhang-etal-2024-llm-based, kim2026rpm}, which appends relevant history snippets retrieved by BM25 (RAP-BM25) or BERTScore (RAP-BERT). BM25 measures lexical overlap, whereas BERTScore captures semantic similarity. (4) Personalized Supervised Fine-Tuning (P-SFT) \cite{tan-etal-2024-democratizing} adapts the model by updating its parameters on personalization data.

\subsection{Experimental Details}
For all non-parametric baselines, we evaluate six instruction-tuned LLMs from four model families, covering a broad range of model scales: Gemma3-27B \cite{gemmateam2025gemma3technicalreport}, Meta-Llama3.1-8B/70B \cite{grattafiori2024llama3herdmodels}, Mistral-Small-3.2-24B\footnote{\url{https://huggingface.co/mistralai/Mistral-Small-3.2-24B-Instruct-2506}}, and Qwen3-14B/235B-A22B \cite{yang2025qwen3technicalreport}.
For parametric fine-tuning, we train six models with LoRA \cite{DBLP:conf/iclr/HuSWALWWC22}, including 
Qwen3-4B/8B/14B \cite{yang2025qwen3technicalreport}, 
Llama3.2-3B/3.1-8B \cite{grattafiori2024llama3herdmodels}, 
and Ministral-8B\footnote{\url{https://mistral.ai/news/ministraux}}.

\subsection{Evaluation Metrics} 
For Task 1, we evaluate personalized generation from two complementary perspectives. (i) We assess whether each generated response realizes the intended personalization strategy. We report accuracy, balanced accuracy, macro-F1, and weighted-F1, which measure overall strategy correctness. (ii) We evaluate response quality by comparing generated responses with the reference responses using standard text generation metrics, including BLEU-4 \cite{10.3115/1073083.1073135}, ROUGE-L \cite{lin-2004-rouge}, METEOR \cite{banerjee-lavie-2005-meteor}, and BERTScore \cite{DBLP:conf/iclr/ZhangKWWA20}. For Task 2, a personalized binary classification task, we report accuracy, precision, recall, F1 score, and Matthews correlation coefficient (MCC).

\subsection{Main Results}
Table~\ref{tab:main_results} reports the main results, averaged over six backbone models for each method.

\noindent \textbf{Overall Trends.} Non-parametric personalization generally improves over Base-LLM, but the gains remain limited. This indicates that user histories contain useful personalization signals, yet prompt augmentation or retrieval alone cannot reliably support personalized reasoning when profile cues and query preferences lie in different conceptual spaces. Among non-parametric methods, PAP-LLM achieves the strongest overall performance, suggesting that summarized long-term profiles provide more stable personalization signals than full-history prompting or local snippet retrieval.

\noindent \textbf{More history does not necessarily improve personalization.} FHP-LLM performs worse than Base-LLM on Task~1 strategy metrics, with Accuracy dropping from 21.48 to 18.36 and Macro-F1 from 14.15 to 10.36. This suggests that directly concatenating long user histories can introduce noise and redundancy, making it difficult for models to identify the personalization cues relevant to the current query. In contrast, PAP-LLM achieves substantially stronger strategy performance. Manual inspection further shows that PAP-LLM often captures task-relevant profile dimensions such as occupation, suggesting that profile summarization can help filter irrelevant historical information.

\noindent \textbf{Semantic retrieval helps, but remains insufficient.} RAP-BERT slightly outperforms RAP-BM25 overall. For example, on Task~2, RAP-BERT achieves an F1 of 61.59, compared with 56.39 for RAP-BM25, while also showing modest advantages on most Task~1 strategy metrics. This suggests that semantic retrieval is more effective than lexical matching for identifying relevant historical information. However, the gap between the two retrieval methods remains small, and both are clearly weaker than PAP-LLM and P-SFT. Thus, retrieving semantically similar history alone appears insufficient for resolving the cross-concept mappings required by \textsc{Vibe-Bench}.

\noindent \textbf{Parametric personalization shows contrasting behavior across tasks.} On Task~2, P-SFT achieves the strongest performance, substantially outperforming all non-parametric methods, with an F1 of 83.15 and an MCC of 0.68. This demonstrates the advantage of parameter updates for learning profile-to-preference mappings. On Task~1, however, P-SFT exhibits a notable discrepancy: it achieves the strongest response-generation scores, including a BLEU-4 of 9.11 and a BERTScore of 14.84, while performing poorly on strategy metrics, with an Accuracy of 17.97 and a Macro-F1 of 10.46. This suggests that fitting the surface form and local semantics of reference responses does not necessarily translate into stable strategy-level personalized reasoning.

\noindent \textbf{Strategy-level personalization remains challenging.} Across all methods, Task~1 strategy performance remains low, with Accuracy ranging from around 18 to 24 and consistently low Macro-F1 scores. Models therefore appear considerably better at generating reference-like responses than at selecting and realizing an appropriate emotion-regulation strategy conditioned jointly on the user profile and current event. These results highlight the central challenge of \textsc{Vibe-Bench}: effective personalization requires reasoning across concept spaces rather than relying on semantic matching.

\subsection{Ablation Study on Benchmark Design}

\begin{table}[]
\centering
\resizebox{\columnwidth}{!}{%
\begin{tabular}{cccccc}
\Xhline{1pt}
\multicolumn{2}{c}{\multirow{2}{*}{Benchmark Design}} & \multicolumn{2}{c}{Task 1 Acc.} & \multicolumn{2}{c}{Task 2 Acc.} \\
\multicolumn{2}{c}{}                                  & \iconqwen           & \iconllama          & \iconqwen           & \iconllama          \\ \Xhline{1pt}
Parad. 1   & \texttt{R}           & 100.0          & 100.0          & 100.0          & 100.0          \\ \cline{1-2}
Parad. 2                            & \texttt{RR}            & 100.0          & 100.0          & 100.0          & 100.0          \\ \cline{1-2}
Parad. 3 (ours)                         & \texttt{PP-RR}     & 18.75          & 17.97          & 85.94          & 84.38          \\
- w/o P-infer                          & \texttt{P-RR}         & 21.09          & 21.88          & 88.28          & 86.72          \\
- w/o R-infer                          & \texttt{PP-R}         & 45.31          & 57.81          & 89.06          & 85.16          \\
- w/o P\&R-infer                   & \texttt{P-R}  & 57.03          & 67.97          & 94.53          & 91.41          \\ \Xhline{1pt}
\end{tabular}%
}
\caption{Ablation study on benchmark designs for different preference reasoning paradigms.}
\label{tab:ablation}
\end{table}

We conduct ablation studies to answer two questions. First, in \textit{cross-concept implicit preference reasoning}, how large is the performance gap relative to (i) Paradigm~1, \textit{explicit preference extraction}, and (ii) Paradigm~2, \textit{within-concept implicit preference reasoning}? Second, under PRCM, does P-SFT fail because models cannot internalize cross-concept profile-preference mappings (i.e., $p_\ast \xrightarrow{\phi} r_0$), or because semantic mismatch prevents the query from anchoring to relevant profile evidence for profile-side reasoning (i.e., $p_0 \rightarrow p_\ast$)? To probe these questions, we extend \textsc{Vibe-Bench} with five information-injection variants.

Paradigm~1 (\texttt{R}) inserts an explicit, query-aligned preference statement, allowing direct retrieval of the answer (e.g., “\textit{I prefer to regulate my emotions via reappraisal}”). Paradigm~2 (\texttt{RR}) supplies explicit profile information and requires a single-step preference inference, e.g., given an occupation-derived RIASEC type, the model judges if the dialogue's activity cues align with this profile. For PRCM, we ablate two reasoning chains: profile inference (\texttt{P-RR}) and preference inference (\texttt{PP-R}). Providing both (\texttt{P-R}) yields the most direct test of if PLLMs can perform cross-concept mapping.

\noindent\textbf{Analysis of Results.} P-SFT achieves perfect accuracy under Paradigms~1 and~2 (Table~\ref{tab:ablation}), indicating that it suffices for conventional benchmarks. In PRCM, however, controlled comparisons reveal a different failure mode. When cross-concept mapping is required (\texttt{R} vs.\ \texttt{P-R}; \texttt{RR} vs.\ \texttt{P-RR}), accuracy drops sharply (58\% on Task~1; 10\% on Task~2), indicating that P-SFT does not reliably induce the abstract profile-preference associations needed for cross-concept transfer. In contrast, ablating the profile-inference chain that is semantically unrelated to the query (\texttt{P-R} vs.\ \texttt{PP-R}; \texttt{P-RR} vs.\ \texttt{PP-RR}) causes relatively minor degradation (7\% and 4\%), showing that semantic misalignment only weakly disrupts anchoring the query to relevant profile evidence for reasoning. Overall, \textbf{PRCM is primarily bottlenecked by cross-concept mapping} itself, while semantic mismatch in anchoring contributes only marginal additional error.

\begin{figure}[!t]
  \centering
  \includegraphics[width=0.85\columnwidth]{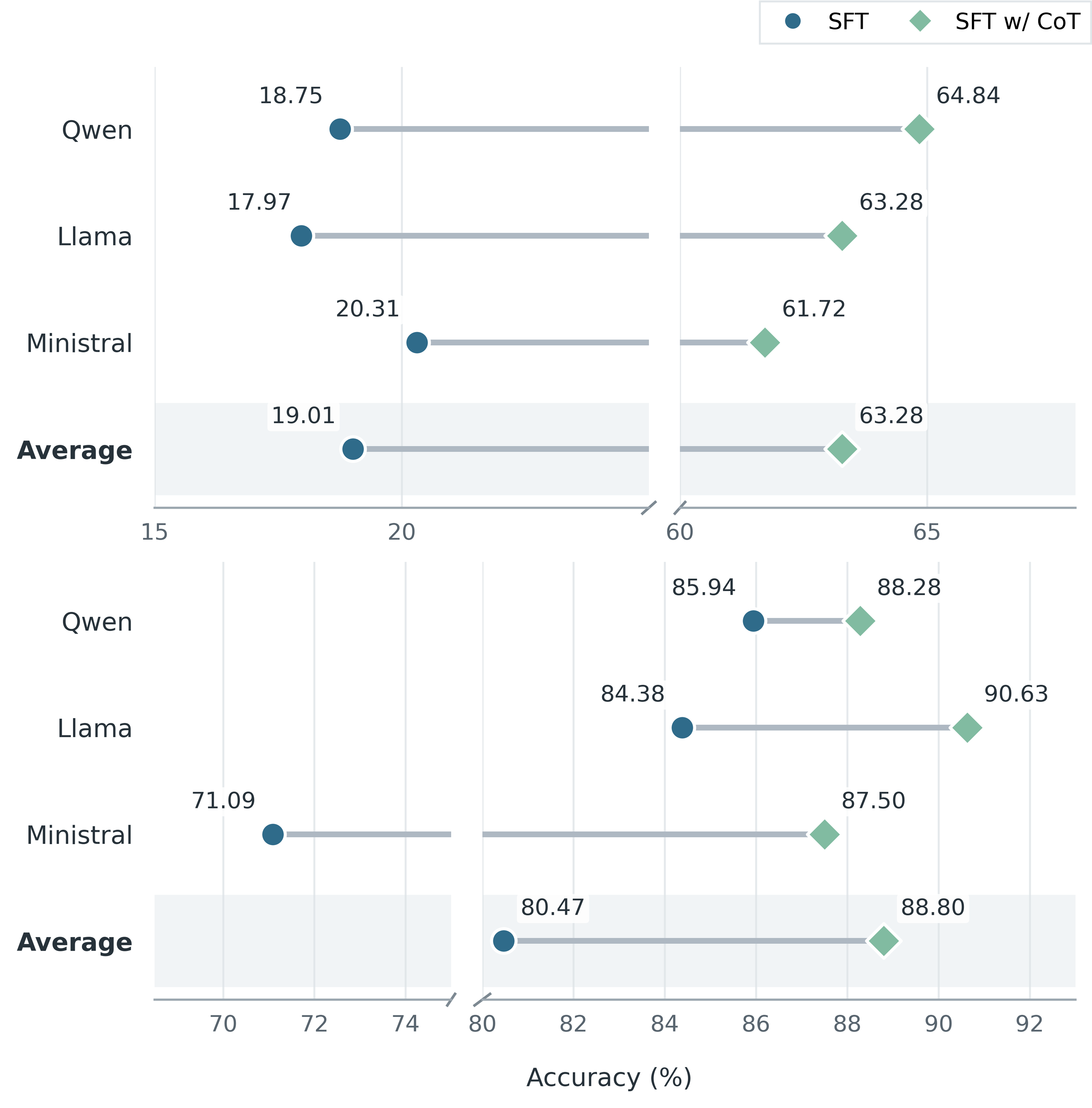}
  \caption{Concept-Aware Reasoning, exemplified by CoT, improves accuracy. \textit{Top}: Task 1; \textit{Bottom}: Task 2.
}
  \label{fig:cot}
\end{figure}

\section{Discussion}

Following Section~\ref{sec:3}, our main experiments adopt a weakly supervised setting where models observe only user histories, current queries, and end-task supervision. Models must therefore identify task-relevant concepts, infer user attributes along these dimensions, and induce the corresponding mappings. We adopt this setting because interaction data are scalable to collect, whereas concept and mapping annotations are costly, domain-specific, and often unavailable. In open-domain personalization, the relevant concept space may be unknown in advance. Under this setting, we evaluate five scalable, general-purpose baselines in Table~\ref{tab:main_results}.

When the relevant concept ontology and cross-concept knowledge are explicitly available, however, \textbf{Concept-Aware Reasoning} substantially improves performance. We therefore conduct an additional experiment using persona-card labels to construct template-based Chain-of-Thought (CoT) rationales \cite{wei2022chain} that guide models through concept inference and cross-concept mapping before answering (Figure~\ref{fig:cot}). This improves accuracy by 44\% on Task 1 and 8\% on Task 2, suggesting that the main challenge under PRCM lies in discovering and inducing the required mappings, rather than merely applying them once provided.

\begin{figure}[!t]
  \centering
  \includegraphics[width=0.65\columnwidth]{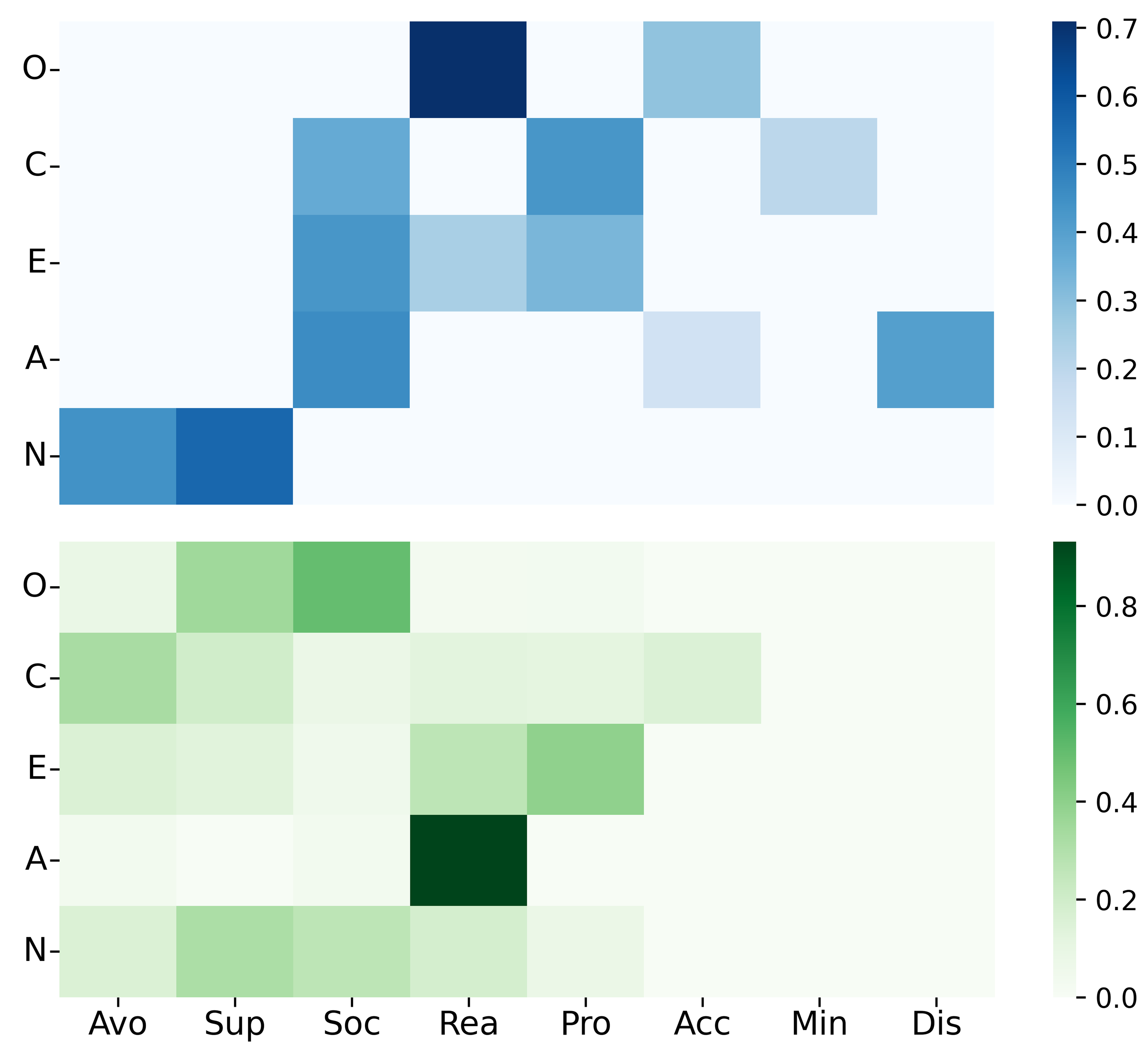}
  \caption{Big Five-emotion-regulation strategy mappings. \textit{Blue}: dataset ground-truth distribution; \textit{Green}: zero-shot induced distribution by \texttt{GPT-5-mini}.}
  \label{fig:heatmap}
\end{figure}

Explicit CoT traces are hard to obtain, so almost all existing benchmarks \cite{liu2025surveypersonalizedlargelanguage} do not include them. This naturally raises a key question for \textsc{Vibe-Bench}: \textbf{Can we automatically discover profile-preference mapping knowledge} from such datasets alone?

We run a preliminary study by supplying full user-specific data, query-response pairs, and a manually defined concept set. For each instance, we prompt the model to assign concept categories and then aggregate label frequencies to extract the induced mapping regularities. Figure~\ref{fig:heatmap} shows that the resulting zero-shot mappings deviate markedly from those implicit in the dataset, indicating strong dependence on both the model's priors and the hand-crafted concept space, along with non-trivial reasoning overhead.

\section{Conclusion and Future Work}

We identify PRCM as a distinct preference-reasoning regime and introduce a taxonomy and benchmark, \textsc{Vibe-Bench}, for evaluating cross-concept personalization beyond semantic retrieval. Experiments show that current personalization methods largely exploit semantic correlations but struggle to induce robust profile-preference mappings, revealing cross-concept mapping as the central bottleneck under PRCM.

Future work should develop training frameworks that can both induce profile-preference mappings from data and robustly internalize them into model parameters, enabling interpretability at both the instance and dataset levels. \textsc{Vibe-Bench} provides a standardized testbed to drive progress on PRCM and personalized reasoning.

\section*{Limitations}

\textsc{Vibe-Bench} is a controlled diagnostic benchmark designed to isolate profile–preference conceptual misalignment, rather than a comprehensive evaluation of real-world personalization. It currently covers two psychology-grounded tasks, uses synthetically generated interaction histories, and is evaluated only in English. Although the construction is grounded in external resources and the gold test set is manually verified, these procedures primarily establish construction quality rather than ecological validity. Whether the observed PRCM failure patterns transfer to naturally occurring user histories, multilingual and cross-cultural settings, additional domains, or real-user outcomes remains an open question.

Our profile–preference mappings should also be interpreted cautiously. They encode probabilistic and context-dependent population-level regularities rather than deterministic prescriptions or individual-level ground truth. Users with similar inferred attributes may still exhibit heterogeneous preferences, and the strength or direction of these associations may vary across contexts and cultures. We therefore view such mappings as weak priors for cold-start-like settings where users have neither explicitly stated the relevant preference nor provided semantically related evidence. They should not override direct user preferences or feedback, and overgeneralizing them to individuals may lead to stereotyped or overly confident personalization.

Our experimental coverage is also limited. The main comparisons focus on scalable prompting, retrieval, and supervised fine-tuning, and we do not evaluate preference-optimization approaches, including reinforcement-learning-based or direct preference-optimization methods. Moreover, while our Concept-Aware Reasoning experiments show that explicit conceptual supervision can substantially improve performance, they rely on predefined concept information, and our study of automatic mapping discovery remains preliminary. Developing methods that automatically discover relevant concepts, induce uncertain cross-concept mappings from interaction data, and robustly internalize them through reasoning or preference optimization is an important direction for future work.

\section*{Ethical Considerations}

\paragraph{I. Intended Use.} \textsc{Vibe-Bench} adopts Holland's RIASEC framework and emotion-regulation strategies solely as test scenarios to study profile-preference conceptual misalignment, and the dataset is not intended for real-world deployment. While both scenarios are grounded in established psychological theories, all training and dialogue data are synthetically generated by language models (with publicly available datasets used as part of the prompting) and may contain biases or unreliable inferences. We therefore restrict the dataset to research use only and prohibit its use for real-world emotion-regulation guidance, mental-health intervention, career planning, or other high-stakes decision-making.

\paragraph{II. Data Privacy, Licensing, and Terms.} \textsc{Vibe-Bench} is entirely synthetic and does not contain real user conversations or personal data. All external datasets, models, and software tools used in this work were accessed and used in accordance with their documented licenses and terms of use.

\paragraph{III. Data and Code Availability.} 
\textsc{Vibe-Bench}, including all benchmark data, evaluation code, and metadata contributed by this work will be made fully available upon publication at \url{https://github.com/yiwenJG/VIBE-Bench}.

\paragraph{IV. Use of AI Tools.}
AI assistants were used to support code development and manuscript editing. The authors carefully reviewed and revised all AI-assisted outputs and take full responsibility for the final content, analyses, and claims in this paper.

\bibliography{custom}

\appendix

\section{Details of Metadata Collection}
\label{appendix:meta}

\subsection{Big Five Personality}
We model personality via two complementary pathways: behavioral cues and psycholinguistic features. Specifically, we collect two publicly available Big Five personality inventories, the \textit{Big Five Inventory} (BFI) \cite{john2010handbook} and the \textit{International Personality Item Pool NEO-300} (IPIP-NEO-300) \cite{goldberg1999broad}, which together comprise 348 self-descriptive sentences covering all five traits. We then manually construct semantically opposite variants of these items to invert their evaluative polarity. For example, \textit{Tell the truth} (a positively keyed item for \textit{Conscientiousness}) is rewritten as \textit{Sometimes tell lies} to characterize lower \textit{Conscientiousness}. This procedure yields 696 behavioral descriptions in total, each corresponding to either a high or a low level of a trait.

In addition, because individual language-use patterns, such as lexical choices, syntactic structure and pragmatic style, systematically reflect Big Five personality traits \cite{vu2025psychadapteradaptingllmtransformers}, we incorporate \textsc{BIG5-CHAT} \cite{li-etal-2025-big5}, a large-scale dataset grounded in psycholinguistic regularities, as metadata to provide language-pattern information associated with each personality profile.

\subsection{Occupational Information Network}
O$^{*}$NET \footnote{\url{https://www.onetonline.org/}}\label{fn:onet} is a publicly available database of occupational information developed by the U.S. Department of Labor. We curate 876 occupation entries spanning 22 major occupational groups from O$^{*}$NET. Each entry includes multiple structured fields: a standardized occupational title; incumbent-reported job titles (on average 8 variants); an occupational summary; required education level; a list of work duties (on average 15 per occupation); and the corresponding interest categories under Holland's RIASEC framework (typically 1-3 types per occupation).

\subsection{Holland's RIASEC Framework}
We include a widely used vocational interest inventory, the \textit{O$^{*}$NET Interest Profiler (Long Form)}\footnotemark[\value{footnote}], which comprises 180 self-descriptive items that ask respondents to indicate their preferences for various activities associated with the RIASEC domains. For instance, \textit{Compose or arrange music} reflects \textit{Artistic} interests, whereas \textit{Teach disabled people work and living skills} reflects \textit{Social} interests.

\subsection{Emotional Distress Event}
We sample 1,142 and 2,362 instances, respectively, from two emotional-support dialogue datasets, ESConv \cite{liu-etal-2021-towards} and ExTES \cite{zheng2023buildingemotionalsupportchatbots}, using stratified sampling over event categories. Each instance includes the description of an emotional distress event and the corresponding multi-turn dialogue. In total, the sampled instances span 48 event themes, such as \textit{Communication Challenges} and \textit{Job Crisis}.

\section{Concept Spaces and Concept-Based Interpretability}
\label{appendix:concept}

\subsection{Concept-Based Interpretability}
Concept-based interpretability explains model behavior through human-interpretable concepts rather than opaque latent features \cite{jiang2025enhancing, jiang2025wise}. A concept represents a meaningful high-level property or category that humans can recognize and reason about. For example, Concept Bottleneck Models explicitly predict such concepts before producing the final task prediction, making the intermediate decision process inspectable and potentially editable \cite{koh2020concept}. Concept-based approaches have also been applied to various domains for interpretable reasoning \cite{mehta2025interpretable,yang2026nerdneurosymbolicruledistillation}.

\subsection{Concept and Concept Space}
In this work, a concept refers to a human-interpretable abstraction that captures a meaningful property, attribute, category, or state. A concept space is a coherent set of concepts that describe a shared semantic or functional domain and can be meaningfully related or compared within that domain. For example, Openness and Extraversion belong to the Big Five personality concept space, whereas Reappraisal and Suppression belong to the emotion-regulation concept space. Here, ``space'' is used in a conceptual rather than geometric sense; it does not necessarily denote a learned embedding or Euclidean vector space.

\subsection{Within- and Cross-Concept Spaces}
Two concepts are considered \textit{within-concept} when they belong to the same or closely related conceptual domain, such that their relationship can be established through direct semantic or conceptual relatedness. In contrast, concepts are \textit{cross-concept} when they belong to distinct human-interpretable domains with little direct semantic or taxonomic overlap. Their relationship therefore cannot generally be recovered from semantic similarity alone and instead requires an additional cross-space mapping, which may be supported by external knowledge, theory, or population-level regularities. For example, a personality trait and an emotion-regulation strategy are semantically distinct concepts, even though psychological evidence may establish systematic associations between them.

\subsection{Definition of Conceptual Alignment}
\label{appendix:ca}
PLLMs operate under \emph{Profile-Preference Conceptual Alignment} (PRCA) if, for each user $u_i$ and query $q^{(i)}_j$, the profile evidence relevant to $q^{(i)}_j$ and the target preference $r^{(i)}_j$ are expressible in a shared concept space. In this paradigm, the model can obtain an initial query-conditioned preference $r_0$ by retrieving and aggregating parts of $p_i$ that are semantically relevant to $q^{(i)}_j$, and, when needed, refine $r_0$ into a query-appropriate preference $r_\ast$ (denoted $r_0 \rightarrow r_\ast$). Most existing benchmarks implicitly assume PRCA: benchmarks with explicit profiles typically treat $r_0$ as the final target (direct extraction), whereas implicit preference reasoning benchmarks evaluate the refinement from $r_0$ to $r_\ast$.

\section{Glossary of Psychological Terms}
\label{sec:scyt}

This section introduces key terminology. These definitions are included in data-synthesis prompts and in the annotation guidelines for human annotators.

\subsection{Big Five Personality Traits}
\label{appendix:bigfive}

The Big Five personality model \citep{mccrae1992introduction} characterizes stable behavioral, cognitive, and affective patterns along five dimensions: Neuroticism, Extraversion, Openness, Agreeableness, and Conscientiousness.

\paragraph{Openness} Openness to experience is a personality dimension that distinguishes imaginative, creative individuals from conventional and down-to-earth ones. It reflects a general appreciation for art, emotion, adventure, unusual ideas, imagination, curiosity, and variety of experience. People high in openness are intellectually curious, sensitive to beauty, and willing to try new things, often thinking and acting in individualistic and nonconforming ways. Compared with closed individuals, they tend to be more aware of their feelings, more creative, and more likely to hold unconventional beliefs. However, open people may sometimes be seen as unpredictable or lacking focus, while those low in openness are more pragmatic, data-driven, and prefer stability and established routines.

\paragraph{Conscientiousness} Conscientiousness refers to a tendency to be self-disciplined, act dutifully, and strive for achievement against external standards or expectations. It reflects how individuals control, regulate, and direct their impulses. People high in conscientiousness prefer planned rather than spontaneous behavior and are often seen as focused and determined, though sometimes stubborn. Those low in conscientiousness tend to be flexible and spontaneous, which can make them appear fun and lively, but may also lead to sloppiness, unreliability, or acting on antisocial impulses.

\paragraph{Extraversion} Extraversion is characterized by a strong engagement with the external world and a preference for breadth of activities and social interaction. Extraverts gain energy from external stimulation and enjoy being with people, often appearing enthusiastic, assertive, and action-oriented. They tend to seek excitement and readily respond to opportunities for activity, frequently taking the lead in group settings and drawing attention through talkativeness and visible energy. In contrast, introverts display lower levels of social engagement and energy, appearing quiet, reserved, and independent, preferring less stimulation and more time alone without being unfriendly or antisocial. Most individuals fall between these two extremes, combining traits of both extraversion and introversion.

\paragraph{Agreeableness} Agreeableness reflects an individual's concern for cooperation and social harmony. People high in agreeableness value getting along with others and tend to be considerate, kind, generous, trusting, and helpful, often willing to compromise their own interests for the sake of others. They generally hold an optimistic view of human nature, believing that people are honest and trustworthy. In contrast, disagreeable individuals prioritize self-interest over social harmony, showing less concern for others' well-being and often appearing suspicious, unfriendly, or uncooperative. Research has shown that agreeableness is positively related to the quality of interpersonal relationships and transformational leadership, suggesting that agreeable people are more effective in fostering positive, cooperative environments, for example, team leaders who motivate through empathy and collaboration rather than authority.

\paragraph{Neuroticism} Neuroticism is the tendency to experience strong and persistent negative emotions such as anger, anxiety, or depression. It reflects emotional instability and low tolerance for stress or change. Individuals high in neuroticism are emotionally reactive, easily upset, and vulnerable to stress. They are more likely to interpret ordinary situations as threatening and perceive minor frustrations as overwhelming. Their negative emotions often last unusually long, leading to frequent bad moods and pessimism toward work or relationships. For example, they may feel anxious about job pressure or dissatisfied with personal achievements, increasing their risk of depression. Neuroticism is associated with problems in emotional regulation, poorer decision-making, and reduced psychological well-being. In contrast, people low in neuroticism are calm, emotionally stable, and less likely to experience persistent negative feelings.

\subsection{Holland Code (RIASEC)}

Holland's RIASEC framework \citep{holland1997making} is a theory of vocational interest-occupation fit that characterizes which types of occupations and work environments best match an individual's interests. It groups vocational interests into six types: Realistic, Investigative, Artistic, Social, Enterprising, and Conventional.

\paragraph{Realistic} Realistic interest refer to a preference for practical, hands-on activities that involve the explicit, ordered, or systematic manipulation of objects, tools, machines, and animals. People with realistic interests enjoy dealing with real-world materials such as wood, plants, or machinery, and often take pleasure in outdoor or physical work. These behavioral tendencies lead to the development of manual, mechanical, agricultural, electrical, and technical skills. They typically dislike occupations that mainly involve paperwork or close social interaction.

\paragraph{Investigative} Investigative interest refers to a preference for activities involving the observation, analysis, and creative investigation of physical, biological, or cultural phenomena to understand and explain them. People with investigative interests enjoy working with ideas more than with physical activity, often searching for facts, solving problems mentally, and developing scientific and mathematical skills. Rather than persuading or leading others, they focus on exploring questions and uncovering underlying principles-for example, conducting experiments, analyzing data, or studying natural or social systems.

\paragraph{Artistic} Artistic Interest refers to a preference for free, ambiguous, and unsystematic activities that involve manipulating physical, verbal, or human materials to create artistic forms or products. People with artistic interests enjoy work related to the artistic side of things-such as forms, designs, and patterns-and value self-expression. They prefer settings that allow creativity and flexibility, where work can be done without following a clear set of rules.

\paragraph{Social} Social Interest refers to a preference for activities that involve assisting, teaching, advising, helping, or otherwise serving others to promote learning, development, or personal growth. Individuals with strong social interests enjoy communicating with people more than working with objects, machines, or data. They tend to engage in roles that inform, train, cure, or enlighten others, such as teaching, counseling, or healthcare work.

\paragraph{Enterprising} Enterprising Interest refers to a preference for activities that involve persuading and leading others to achieve organizational goals or economic gain. People with this interest enjoy starting up and carrying out projects, especially business ventures. They like making decisions, taking risks for profit, and prefer action rather than abstract thinking. Such behavioral tendencies often lead to the development of leadership, interpersonal, and persuasive competencies.

\paragraph{Conventional} Conventional Interest refers to a preference for activities that involve the explicit, ordered, and systematic manipulation of data, such as keeping records, filing materials, reproducing documents, and organizing business machines or data processing equipment to achieve organizational or economic goals. People with conventional interests like work that follows set procedures and routines, where there are clear standards and lines of authority. They prefer working with data and details rather than ideas, and tend to acquire clerical, computational, and business system competencies.

\subsection{Emotion-Regulation Strategies}

Emotion regulation is the process by which individuals modulate the experience and expression of their emotional states through diverse cognitive and behavioural strategies \cite{gross1998emerging,gratz2004multidimensional,pena2015integrating}. In this study, we select eight emotion-regulation strategies: Avoidance, Suppression, Social Support, Reappraisal, Problem Solving, Acceptance, Mindfulness and Distraction.

\paragraph{Avoidance} Avoidance refers to efforts to distance oneself from distressing emotions, thoughts, memories, or external situations that might trigger negative affect. This strategy can include both behavioural avoidance, such as steering clear of specific people or contexts, and internal avoidance, such as trying not to think about or feel unpleasant psychological experiences.

\paragraph{Suppression} Suppression refers to a conscious and deliberate effort to push unwanted thoughts out of awareness.

\paragraph{Social Support} Social support involves turning to others, such as friends, family, or close contacts for emotional or practical help during times of stress. This strategy draws on interpersonal resources like encouragement, humour, or assistance, forming a support network that can aid emotion regulation.

\paragraph{Reappraisal} Reappraisal is reframing how one interprets a situation in order to change its emotional impact. This may include altering the perceived meaning of the event or its relevance to the self, often by generating more positive or neutral interpretations.

\paragraph{Problem Solving} Problem solving involves deliberate cognitive and behavioural efforts to directly address and change a distressing situation. It involves identifying stressors, planning effective actions, and implementing strategies to reduce or eliminate the source of emotional discomfort.

\paragraph{Acceptance} Acceptance is the acknowledgment and allowance of difficult thoughts and emotions without judgment, avoidance, or attempts to control or change them. It reflects a present-focused, non-evaluative stance toward internal experiences, fostering openness and psychological flexibility.

\paragraph{Mindfulness} Mindfulness is cultivating a nonjudgmental, open awareness of one's present-moment experiences, including thoughts and emotions. It reflects a receptive attentiveness to internal and external events as they unfold, without trying to change or evaluate them.

\paragraph{Distraction} Distraction is a strategy that gently redirects the user's attention away from distressing thoughts or feelings by engaging their senses, shifting focus, or introducing simple cognitive tasks to ease immediate distress.

\section{Concept-Mapping Distribution}

Table~\ref{tab:personality-strategy} summarizes the distribution of emotion-regulation strategies mapped to Big Five traits in \textsc{Vibe-Bench}. Table~\ref{tab:job-riasec} reports the probabilistic mapping from O$^{*}$NET major occupational groups to RIASEC types. Importantly, Table~\ref{tab:personality-strategy} is a deliberately human-designed mapping that is clean and unambiguous, whereas Table~\ref{tab:job-riasec} is only defined at the coarse group level. Since O$^{*}$NET spans 876 specific occupations whose duties can correspond to different RIASEC types, a major occupational group cannot fully capture the RIASEC profile implied by the underlying job responsibilities.

\section{More Data Distribution}

We provide additional post-split distribution statistics in Tables~\ref{tab:bigfive-dis},~\ref{tab:riasec-dis}~and~\ref{tab:job-dis}.

\section{Human Annotation}

Three PhD students with different academic backgrounds (psychology, computer science, and education) voluntarily participated in manually annotating 128 test instances. To minimize potential bias, we did not disclose the dataset construction rationale to the annotators. They were provided only with the definitions of the relevant psychological terms and a reference list of O$^{*}$NET major occupational groups along with the corresponding specific occupation titles. No identifiable or personal information was provided at any stage.

\section{Implementation Details}

We fine-tune all LLMs using LLaMA-Factory \cite{zheng2024llamafactoryunifiedefficientfinetuning} with LoRA \cite{DBLP:conf/iclr/HuSWALWWC22}. We use a LoRA rank of 8, a batch size of 8, a learning rate of 1e-4, and train for 5 epochs. We apply a cosine learning-rate schedule with a warmup ratio of 0.1 and enable bfloat16 (bf16) mixed-precision training. We select the checkpoint with the best validation performance and report its corresponding results on the test set. Experiments are conducted on four RTX A5000 GPUs, and each fine-tuning run takes approximately 3 hours.

PAP-LLM summaries are generated using a designed prompt that instructs the LLM to summarize the user profile from the interaction history. Consistent with our experimental setting, we do not specify predefined concepts; instead, the model independently identifies and summarizes the relevant profile information.

\section{Response Quality Assessment Rubric}
\label{appendix:rubric}

\subsection{Event Grounding}
\paragraph{Definition} This dimension evaluates whether the response is grounded in the user’s specific emotional distress event, including the situation, people involved, emotions, concerns, and contextual details. Do not reward a response merely for being empathetic. Reward it only if the empathy and guidance are specifically connected to the user’s described event.

1: The response is generic or unrelated to the user’s event. It could apply to almost any emotional situation and does not mention or reflect the specific distress described by the user.

2: The response shows minimal connection to the event, such as acknowledging a broad emotion, but misses most key details. It may feel template-like or only loosely relevant.

3: The response correctly identifies the general situation and emotional state, but only uses limited event details. It is relevant, but not strongly tailored to the user’s specific experience.

4: The response clearly reflects the user’s event, emotions, and main concern. It uses several concrete details from the query and gives advice that fits the described situation. Minor details may still be missing.

5: The response is deeply grounded in the user’s specific event. It accurately incorporates the situation, emotions, interpersonal/contextual factors, and the user’s expressed concern. The response feels clearly written for this user and this event, not reusable as a generic answer.

\subsection{Helpfulness}

\paragraph{Definition} This dimension evaluates whether the response provides supportive, concrete, and usable guidance that could reasonably help the user manage their distress in the controlled benchmark setting. Do not give a high helpfulness score just because the response is long. A concise response can score high if it is specific, supportive, and actionable.

1: The response is unhelpful, dismissive, confusing, or lacks meaningful support. It provides no clear guidance for the user.

2: The response contains some supportive language but little practical value. Advice is vague, unrealistic, overly general, or difficult to apply.

3: The response is moderately helpful. It offers relevant support or advice, but the guidance may be somewhat generic, incomplete, or not clearly actionable.

4: The response is helpful and practically useful. It provides clear emotional support and realistic guidance that the user could apply. It is mostly concrete and well matched to the user’s situation.

5: The response is highly helpful. It combines emotional validation with clear, specific, realistic, and easy-to-follow guidance. The user would likely know what to do next after reading it.

\subsection{Psychological Appropriateness}

\paragraph{Definition} This dimension evaluates whether the response is psychologically appropriate, emotionally safe, and within the scope of non-clinical emotion-regulation guidance. It should avoid harmful, dismissive, judgmental, overconfident, or clinically inappropriate content. A response should not be penalized simply because it does not provide professional therapy. The key question is whether it is safe and appropriate as a controlled emotion-regulation reference response.

1: The response is clearly inappropriate or unsafe. It may blame the user, minimize their distress, encourage harmful behavior, make unsupported clinical claims, or give advice that could worsen the situation.

2: The response has noticeable psychological appropriateness issues. It may be overly directive, invalidating, insensitive, too casual for the distress level, or may overstep into diagnosis or therapy-like claims.

3: The response is generally safe but has limitations. It is not harmful, but may be somewhat shallow, mildly invalidating, overly certain, or insufficiently careful with emotional distress.

4: The response is psychologically appropriate and safe. It validates the user’s feelings, avoids blame and diagnosis, stays within non-clinical support, and gives careful guidance. Minor wording issues may remain.

5: The response is highly appropriate and safe. It is warm, respectful, nonjudgmental, emotionally sensitive, and carefully scoped. It avoids overpromising, diagnosis, clinical intervention, or pressure, while still providing meaningful support.

\subsection{Strategy Realization}

\paragraph{Definition} This dimension evaluates whether the response clearly and coherently realizes one identifiable emotion-regulation strategy. The evaluator is given the strategy taxonomy but is not told which strategy was intended for the response. A high score indicates that the response makes one dominant strategy recognizable through concrete guidance and consistent regulatory mechanisms.

1: The response does not realize any recognizable emotion-regulation strategy. It is generic, incoherent, purely empathetic, or unrelated to emotion regulation. The evaluator cannot reasonably infer which strategy, if any, the response is trying to implement.

2: The response contains weak traces of one or more strategies, but the strategy is unclear or highly ambiguous. It may provide broad emotional support, but it does not operationalize a clear regulatory mechanism.

3: The response appears to reflect a recognizable strategy, but the implementation is basic, incomplete, or mixed with other strategies in a way that reduces clarity. The evaluator can infer a likely strategy, but with only moderate confidence.

4: The response clearly realizes one dominant emotion-regulation strategy. The guidance is mostly consistent with that strategy’s definition and mechanism, and the evaluator can identify the strategy with high confidence. Minor omissions or mild blending with other strategies are acceptable.

5: The response strongly and unambiguously realizes one dominant strategy. It provides concrete, coherent, and strategy-specific guidance, uses the user’s event as the context for applying the strategy, and makes the regulatory mechanism clear. The evaluator can identify the intended strategy with very high confidence.

\subsection{Conversational Naturalness}

\paragraph{Definition} This dimension evaluates whether the response is fluent, coherent, and human-like in tone and structure. A natural response should read like a supportive conversational reply rather than a mechanical checklist, an encyclopedic explanation, or a rigid list of disconnected options. Do not assign a high score solely because the response is grammatically correct. Penalize responses that are overly list-like, encyclopedic, mechanically structured, or that present many loosely connected options without a natural conversational flow.

1: The response is highly unnatural, robotic, fragmented, or difficult to read. It may be dominated by rigid templates, excessive bullet points, disconnected instructions, or encyclopedia-like explanations.

2: The response is understandable but noticeably mechanical. It may rely heavily on list-like advice, repeated sentence patterns, or too many alternative options, making it feel more like a checklist than a supportive conversation.

3: The response is generally fluent and coherent, but still somewhat formulaic or overly structured. It reads acceptably, but parts may feel generic, instructional, or less conversational.

4: The response is fluent, coherent, and mostly natural. It uses a supportive conversational tone, presents guidance smoothly, and avoids excessive listing or mechanical structure. Minor stiffness is acceptable.

5: The response is highly natural, human-like, and conversational. It flows smoothly, uses warm and context-appropriate language, integrates guidance naturally, and feels like a thoughtful supportive reply rather than a scripted or encyclopedic answer.

\section{Evaluation Method for Strategy Entailment in Generation}

For non-parametric methods, we prompt the instruction-tuned models to generate responses in JSON format with two fields: the adopted strategy and the response content. We then evaluate the strategy explicitly output by the model.

For parametric methods, since the models are fine-tuned to fit the training-data distribution, we train a classifier on the training set to identify the strategy entailed in each generated textual response. The classifier is trained based on Qwen3. It achieves 100\% classification performance on the test set. In parallel, human experts annotate the strategies entailed in the test-set responses, achieving an agreement of 99.31\%. These results demonstrate the reliability of the classifier and its strong consistency with human expert annotations.

\section{Generation Artifact Diagnostic}
\label{appendix:generation-artifacts}

Since \textsc{Vibe-Bench} is synthetically generated, a potential concern is that models may exploit unintended generation artifacts, such as recurring lexical, stylistic, or template-level patterns, rather than learning meaningful relationships between user profiles and target preferences. We conduct a diagnostic experiment to examine this possibility.

Specifically, we randomly shuffle the user histories across instances while keeping the Task~2 labels unchanged, thereby breaking the correspondence between profile evidence and the target label while preserving the marginal distributions and generation characteristics of the dataset. We then fine-tune Qwen3-4B and Llama3.1-8B on the shuffled training data using the same supervised fine-tuning paradigm. Results are reported in Table~\ref{tab:generation-artifact}.

\begin{table}[t]
\centering
\small
\setlength{\tabcolsep}{4.5pt}
\renewcommand{\arraystretch}{1.08}
\begin{tabular}{lccccc}
\toprule
\textbf{Model} & \textbf{Acc.} & \textbf{Prec.} & \textbf{Rec.} & \textbf{F1} & \textbf{MCC} \\
\midrule
Qwen3-4B                 & 81 & 86 & 75 & 80 & 0.63 \\
Qwen3-4B (Shuffled)      & 52 & 51 & 92 & 65 & 0.05 \\
Llama3.1-8B              & 85 & 92 & 77 & 84 & 0.71 \\
Llama3.1-8B (Shuffled)   & 48 & 49 & 64 & 55 & -0.03 \\
\bottomrule
\end{tabular}
\caption{Diagnostic experiment for potential generation artifacts on Task~2. 
\textit{Shuffled} denotes randomly permuting user histories across instances.}
\label{tab:generation-artifact}
\end{table}

After shuffling the user histories, both fine-tuned models degrade to approximately chance-level accuracy, with MCC scores of $0.05$ and $-0.03$, respectively. The relatively high recall under shuffling reflects a degenerate prediction bias and is therefore not indicative of meaningful classification ability; MCC, which accounts for all entries of the confusion matrix, remains close to 0. These results indicate that once the correspondence between user histories and labels is removed, the models cannot recover useful predictive signals from residual generation patterns alone. This provides evidence that \textsc{Vibe-Bench} contains limited exploitable generation artifacts that could serve as shortcuts.

\section{Details of Semantic Similarity Analysis}
\label{appendix:semantic_similarity}

Figure~\ref{fig:sim} reports Query-History and Answer-History semantic similarity computed using BERTScore. As each user history contains multiple utterances, we compute the similarity between the query/answer and each historical user utterance and use the maximum score for each instance. This evaluates whether any historical evidence can be directly retrieved through semantic matching. The low similarity observed for \textsc{Vibe-Bench} supports its intended cross-concept design.

\begin{table*}[]
\centering
\resizebox{0.92\textwidth}{!}{%
\begin{tabular}{cccccc}
\Xhline{1pt}
                & Neuroticism & Extraversion & Openness & Agreeableness & Conscientiousness \\ \Xhline{1pt}
Reappraisal     & -           & 172          & 485      & -             & -                 \\
Problem Solving & -           & 234          & -        & -             & 299               \\
Mindfulness     & -           & -            & -        & -             & 140               \\
Acceptance      & -           & -            & 199      & 97            & -                 \\
Social Support  & -           & 306          & -        & 322           & 255               \\
Avoidance       & 316         & -            & -        & -             & -                 \\
Suppression     & 397         & -            & -        & -             & -                 \\
Distraction     & -           & -            & -        & 282           & -                 \\ \Xhline{1pt}
\end{tabular}%
}
\caption{Emotion-regulation strategy-to-Big Five mapping distributions (occurrence counts) in \textsc{Vibe-Bench}.}
\label{tab:personality-strategy}
\end{table*}

\begin{table*}[]
\centering
\resizebox{0.95\textwidth}{!}{%
\begin{tabular}{ccccccc}
\Xhline{1pt}
                                               & R & I & A & S & E & C \\ \Xhline{1pt}
Management                                     & 9.35      & 7.19          & 0.72     & 10.79  & 36.69        & 35.25        \\
Business and Financial Operations              & 8.26      & 17.36         & -        & 8.26   & 30.58        & 35.54        \\
Computer and Mathematical                      & 14.29     & 38.96         & 1.30     & 1.30   & 6.49         & 37.66        \\
Architecture and Engineering                   & 33.77     & 33.11         & 1.32     & -      & -            & 31.79        \\
Life, Physical, and Social Science             & 26.80     & 36.60         & 0.65     & 5.23   & 3.92         & 26.80        \\
Community and Social Service                   & -         & 20.69         & -        & 48.28  & 17.24        & 13.79        \\
Legal                                          & -         & 20.00         & -        & 10.00  & 35.00        & 35.00        \\
Educational Instruction and Library            & 3.17      & 31.75         & 7.94     & 46.03  & 1.59         & 9.52         \\
Arts, Design, Entertainment, Sports, and Media & 20.00     & 3.33          & 33.33    & 10.00  & 18.89        & 14.44        \\
Healthcare Practitioners and Technical         & 24.24     & 32.47         & 0.87     & 25.97  & -            & 16.45        \\
Healthcare Support                             & 28.07     & 14.04         & -        & 24.56  & 1.75         & 31.58        \\
Protective Service                             & 31.82     & 10.61         & -        & 6.06   & 15.15        & 36.36        \\
Food Preparation and Serving Related           & 31.11     & -             & -        & 13.33  & 20.00        & 35.56        \\
Building and Grounds Cleaning and Maintenance  & 44.44     & -             & -        & -      & 11.11        & 44.44        \\
Personal Care and Service                      & 24.36     & -             & 5.13     & 26.92  & 15.38        & 28.21        \\
Sales and Related                              & 7.41      & -             & 3.70     & 12.96  & 38.89        & 37.04        \\
Office and Administrative Support              & 12.07     & 0.86          & 1.72     & 16.38  & 25.86        & 43.10        \\
Farming, Fishing, and Forestry                 & 48.00     & 4.00          & -        & -      & 4.00         & 44.00        \\
Construction and Extraction                    & 47.29     & 3.88          & 0.78     & -      & 1.55         & 46.51        \\
Installation, Maintenance, and Repair          & 41.67     & 15.00         & 0.83     & -      & 0.83         & 41.67        \\
Production                                     & 44.49     & 5.08          & 5.93     & -      & 0.85         & 43.64        \\
Transportation and Material Moving             & 39.83     & 3.39          & -        & 5.08   & 11.86        & 39.83        \\ \Xhline{1pt}
\end{tabular}%
}
\caption{Probability (\%) distribution of the mapping between major occupational categories and RIASEC in O$^{*}$NET.}
\label{tab:job-riasec}
\end{table*}

\begin{table*}[]
\centering
\resizebox{0.78\textwidth}{!}{%
\begin{tabular}{ccccccccc}
\Xhline{1pt}
\multirow{2}{*}{Big Five} & \multicolumn{2}{c}{Train} & \multicolumn{2}{c}{Valid} & \multicolumn{2}{c}{Test} & \multicolumn{2}{c}{Overall} \\
                          & High        & Low         & High        & Low         & High        & Low        & High         & Low          \\ \Xhline{1pt}
Openness                  & 496         & 623         & 162         & 231         & 26          & 25         & 684          & 879          \\
Conscientiousness         & 512         & 673         & 158         & 245         & 24          & 36         & 694          & 954          \\
Extraversion              & 508         & 665         & 178         & 206         & 26          & 34         & 712          & 905          \\
Agreeableness             & 507         & 634         & 168         & 214         & 26          & 33         & 701          & 881          \\
Neuroticism               & 509         & 672         & 178         & 198         & 26          & 33         & 713          & 903          \\ \hline
Overall                   & 2532        & 3267        & 844         & 1094        & 128         & 161        & 3504         & 4522         \\ \Xhline{1pt}
\end{tabular}%
}
\caption{Distribution over the Big Five traits after dataset splitting, counted by the number of dialogues.}
\label{tab:bigfive-dis}
\end{table*}

\begin{table*}[]
\centering
\resizebox{0.45\textwidth}{!}{%
\begin{tabular}{ccccc}
\Xhline{1pt}
RIASEC        & Train & Valid & Test & Overall \\ \Xhline{1pt}
Realistic     & 474   & 143   & 21   & 638     \\
Investigative & 412   & 125   & 21   & 558     \\
Artistic      & 378   & 134   & 22   & 534     \\
Social        & 424   & 124   & 21   & 569     \\
Enterprising  & 376   & 145   & 21   & 542     \\
Conventional  & 468   & 173   & 22   & 663     \\ \hline
Overall       & 2532  & 844   & 128  & 3504    \\ \Xhline{1pt}
\end{tabular}%
}
\caption{Distribution over RIASEC types, counted by persona cards.}
\label{tab:riasec-dis}
\end{table*}

\begin{table*}[]
\centering
\resizebox{0.8\textwidth}{!}{%
\begin{tabular}{ccccc}
\Xhline{1pt}
Occupation                                     & Train & Valid & Test & Overall \\ \Xhline{1pt}
Management                                     & 153   & 51    & 4    & 208     \\
Business and Financial Operations              & 126   & 42    & 4    & 172     \\
Computer and Mathematical                      & 87    & 29    & 4    & 120     \\
Architecture and Engineering                   & 150   & 50    & 4    & 204     \\
Life, Physical, and Social Science             & 162   & 54    & 8    & 224     \\
Community and Social Service                   & 39    & 13    & 4    & 56      \\
Legal                                          & 18    & 6     & 4    & 28      \\
Educational Instruction and Library            & 177   & 59    & 8    & 244     \\
Arts, Design, Entertainment, Sports, and Media & 105   & 35    & 8    & 148     \\
Healthcare Practitioners and Technical         & 243   & 81    & 4    & 328     \\
Healthcare Support                             & 54    & 18    & 4    & 76      \\
Protective Service                             & 69    & 23    & 8    & 100     \\
Food Preparation and Serving Related           & 45    & 15    & 4    & 64      \\
Building and Grounds Cleaning and Maintenance  & 21    & 7     & 4    & 32      \\
Personal Care and Service                      & 81    & 27    & 8    & 116     \\
Sales and Related                              & 57    & 19    & 8    & 84      \\
Office and Administrative Support              & 144   & 48    & 8    & 200     \\
Farming, Fishing, and Forestry                 & 33    & 11    & 4    & 48      \\
Construction and Extraction                    & 177   & 59    & 8    & 244     \\
Installation, Maintenance, and Repair          & 147   & 49    & 4    & 200     \\
Production                                     & 309   & 103   & 8    & 420     \\
Transportation and Material Moving             & 135   & 45    & 8    & 188     \\ \Xhline{1pt}
\end{tabular}%
}
\caption{Distribution over occupations, counted by persona cards.}
\label{tab:job-dis}
\end{table*}

\clearpage
\onecolumn


\begin{table*}[t]
\centering
\small
\setlength{\tabcolsep}{3pt}
\renewcommand{\arraystretch}{1.08}
\resizebox{\textwidth}{!}{%
\begin{tabular}{llrrrrrrrr}
\toprule
\textbf{Method} & \textbf{Model} 
& \textbf{Acc} 
& \textbf{Balanced Acc} 
& \textbf{Macro-F1} 
& \textbf{Weighted-F1} 
& \textbf{BLEU-4} 
& \textbf{ROUGE-L} 
& \textbf{METEOR} 
& \textbf{BERT-F1} \\
\midrule

\multirow{7}{*}{Base-LLM}
& gemma-3-27b-it & 19.53 & 19.58 & 13.86 & 16.07 & 3.44 & 17.30 & 28.93 & 7.24 \\
& Meta-Llama-3.1-70B-Instruct & 19.53 & 16.41 & 12.63 & 16.36 & 3.05 & 17.36 & 23.66 & 9.84 \\
& Meta-Llama-3.1-8B-Instruct & 20.31 & 21.51 & 14.54 & 16.48 & 2.54 & 16.91 & 19.89 & 8.53 \\
& Mistral-Small-3.2-24B-Instruct-2506 & 22.66 & 20.44 & 17.50 & 20.31 & 3.17 & 17.81 & 22.63 & 9.75 \\
& Qwen3-14B & 21.88 & 17.60 & 11.85 & 15.82 & 2.31 & 17.05 & 23.50 & 9.38 \\
& Qwen3-235B-A22B-Instruct-2507 & 25.00 & 23.33 & 14.52 & 16.42 & 4.09 & 18.33 & 29.76 & 8.33 \\
& \textbf{Average} & 21.48 & 19.81 & 14.15 & 16.91 & 3.10 & 17.46 & 24.73 & 8.84 \\

\midrule
\multirow{7}{*}{FHP-LLM}
& gemma-3-27b-it & 14.84 & 14.48 & 8.55 & 10.38 & 4.15 & 17.98 & 27.54 & 8.89 \\
& Meta-Llama-3.1-70B-Instruct & 16.41 & 15.67 & 7.80 & 9.96 & 1.47 & 15.27 & 17.51 & 8.01 \\
& Meta-Llama-3.1-8B-Instruct & 18.75 & 18.39 & 11.67 & 13.01 & 2.55 & 16.62 & 18.11 & 7.51 \\
& Mistral-Small-3.2-24B-Instruct-2506 & 20.31 & 19.87 & 14.43 & 16.79 & 3.40 & 17.65 & 22.28 & 9.44 \\
& Qwen3-14B & 17.19 & 15.83 & 8.21 & 10.73 & 2.01 & 16.37 & 20.13 & 9.07 \\
& Qwen3-235B-A22B-Instruct-2507 & 22.66 & 18.96 & 11.51 & 13.99 & 4.60 & 18.35 & 25.79 & 7.53 \\
& \textbf{Average} & 18.36 & 17.20 & 10.36 & 12.48 & 3.03 & 17.04 & 21.89 & 8.41 \\

\midrule
\multirow{7}{*}{PAP-LLM}
& gemma-3-27b-it & 25.00 & 22.36 & 19.36 & 22.92 & 3.80 & 17.83 & 30.05 & 7.58 \\
& Meta-Llama-3.1-70B-Instruct & 23.44 & 20.14 & 15.73 & 20.19 & 2.91 & 17.48 & 24.14 & 9.81 \\
& Meta-Llama-3.1-8B-Instruct & 21.88 & 20.83 & 15.83 & 17.78 & 3.17 & 17.72 & 21.50 & 9.74 \\
& Mistral-Small-3.2-24B-Instruct-2506 & 28.12 & 26.42 & 24.95 & 26.84 & 3.75 & 18.11 & 24.14 & 10.80 \\
& Qwen3-14B & 21.09 & 16.74 & 11.69 & 15.41 & 2.73 & 16.79 & 23.27 & 8.14 \\
& Qwen3-235B-A22B-Instruct-2507 & 25.78 & 24.58 & 18.12 & 19.82 & 3.73 & 17.85 & 30.09 & 6.02 \\
& \textbf{Average} & 24.22 & 21.85 & 17.61 & 20.49 & 3.35 & 17.63 & 25.53 & 8.68 \\

\midrule
\multirow{7}{*}{RAP-BM25}
& gemma-3-27b-it & 17.97 & 15.99 & 12.35 & 14.68 & 3.67 & 17.85 & 28.40 & 8.75 \\
& Meta-Llama-3.1-70B-Instruct & 19.53 & 16.26 & 12.81 & 16.82 & 2.55 & 17.15 & 23.29 & 8.93 \\
& Meta-Llama-3.1-8B-Instruct & 19.53 & 18.07 & 14.45 & 16.64 & 2.34 & 16.84 & 18.72 & 8.06 \\
& Mistral-Small-3.2-24B-Instruct-2506 & 24.22 & 20.83 & 18.91 & 22.55 & 2.98 & 17.77 & 21.78 & 10.81 \\
& Qwen3-14B & 22.66 & 18.10 & 12.83 & 17.53 & 2.77 & 16.96 & 22.76 & 9.50 \\
& Qwen3-235B-A22B-Instruct-2507 & 25.00 & 23.65 & 16.63 & 18.45 & 4.56 & 18.50 & 28.81 & 8.57 \\
& \textbf{Average} & 21.48 & 18.82 & 14.66 & 17.78 & 3.15 & 17.51 & 23.96 & 9.10 \\

\midrule
\multirow{7}{*}{RAP-BERT}
& gemma-3-27b-it & 23.44 & 21.51 & 16.65 & 19.17 & 4.34 & 18.11 & 29.34 & 9.44 \\
& Meta-Llama-3.1-70B-Instruct & 24.22 & 20.63 & 15.78 & 20.30 & 2.79 & 17.16 & 23.13 & 9.45 \\
& Meta-Llama-3.1-8B-Instruct & 21.88 & 21.04 & 16.76 & 18.51 & 2.12 & 16.70 & 19.06 & 7.60 \\
& Mistral-Small-3.2-24B-Instruct-2506 & 23.44 & 20.36 & 16.52 & 19.49 & 2.93 & 17.88 & 21.71 & 10.70 \\
& Qwen3-14B & 21.88 & 18.18 & 12.04 & 16.10 & 2.67 & 17.31 & 22.62 & 9.87 \\
& Qwen3-235B-A22B-Instruct-2507 & 21.09 & 19.38 & 14.58 & 16.68 & 3.84 & 18.06 & 28.72 & 7.43 \\
& \textbf{Average} & 22.66 & 20.18 & 15.39 & 18.38 & 3.11 & 17.54 & 24.10 & 9.08 \\

\midrule
\multirow{7}{*}{P-SFT}
& qwen3-4b & 15.62 & 13.99 & 8.10 & 10.51 & 10.22 & 24.59 & 29.31 & 14.26 \\
& qwen3-8b & 21.88 & 21.73 & 16.52 & 17.12 & 11.36 & 26.05 & 31.01 & 16.53 \\
& qwen3-14b & 17.19 & 15.82 & 10.08 & 12.13 & 10.37 & 24.66 & 29.42 & 14.15 \\
& llama3.2-3b & 17.97 & 17.49 & 10.59 & 11.99 & 10.80 & 25.10 & 29.91 & 14.13 \\
& llama3.1-8b & 19.53 & 17.49 & 10.14 & 13.26 & 10.64 & 25.18 & 29.82 & 14.55 \\
& ministral-8b & 15.62 & 14.79 & 7.30 & 8.38 & 1.24 & 34.33 & 20.02 & 15.43 \\
& \textbf{Average} & 17.97 & 16.88 & 10.46 & 12.23 & 9.11 & 26.65 & 28.25 & 14.84 \\

\bottomrule
\end{tabular}%
}
\caption{Main results across different methods and backbone models on Task 1.}
\label{tab:task1_results}
\end{table*}


\begin{table*}[t]
\centering
\small
\setlength{\tabcolsep}{4pt}
\renewcommand{\arraystretch}{1.08}
\resizebox{0.78\textwidth}{!}{%
\begin{tabular}{llccccc}
\toprule
\multirow{2}{*}{\textbf{Method}} 
& \multirow{2}{*}{\textbf{Model}} 
& \multicolumn{5}{c}{\textbf{Task 2}} \\
\cmidrule(lr){3-7}
& & \textbf{Acc.} & \textbf{Prec.} & \textbf{Rec.} & \textbf{F1} & \textbf{MCC} \\
\midrule

\multirow{7}{*}{FHP-LLM}
& \texttt{gemma-3-27b-it} & 65.62 & 68.52 & 57.81 & 62.71 & 0.32 \\
& \texttt{Meta-Llama-3.1-70B-Instruct} & 60.16 & 58.23 & 71.88 & 64.34 & 0.21 \\
& \texttt{Meta-Llama-3.1-8B-Instruct} & 51.56 & 50.88 & 90.62 & 65.17 & 0.05 \\
& \texttt{Mistral-Small-3.2-24B-Instruct-2506} & 69.53 & 69.84 & 68.75 & 69.29 & 0.39 \\
& \texttt{Qwen3-14B} & 60.16 & 67.57 & 39.06 & 49.50 & 0.22 \\
& \texttt{Qwen3-235B-A22B-Instruct-2507} & 57.81 & 55.43 & 79.69 & 65.38 & 0.17 \\
& \textbf{Average} & \textbf{60.81} & \textbf{61.74} & \textbf{67.97} & \textbf{62.73} & \textbf{0.23} \\
\midrule

\multirow{7}{*}{PAP-LLM}
& \texttt{gemma-3-27b-it} & 60.16 & 58.02 & 73.44 & 64.83 & 0.21 \\
& \texttt{Meta-Llama-3.1-70B-Instruct} & 67.19 & 64.86 & 75.00 & 69.57 & 0.35 \\
& \texttt{Meta-Llama-3.1-8B-Instruct} & 64.06 & 59.78 & 85.94 & 70.51 & 0.31 \\
& \texttt{Mistral-Small-3.2-24B-Instruct-2506} & 61.72 & 62.30 & 59.38 & 60.80 & 0.23 \\
& \texttt{Qwen3-14B} & 58.59 & 57.53 & 65.62 & 61.31 & 0.17 \\
& \texttt{Qwen3-235B-A22B-Instruct-2507} & 61.72 & 56.76 & 98.44 & 72.00 & 0.35 \\
& \textbf{Average} & \textbf{62.24} & \textbf{59.88} & \textbf{76.30} & \textbf{66.50} & \textbf{0.27} \\
\midrule

\multirow{7}{*}{RAP-BM25}
& \texttt{gemma-3-27b-it} & 53.91 & 54.55 & 46.88 & 50.42 & 0.08 \\
& \texttt{Meta-Llama-3.1-70B-Instruct} & 57.03 & 58.82 & 46.88 & 52.17 & 0.14 \\
& \texttt{Meta-Llama-3.1-8B-Instruct} & 53.91 & 52.48 & 82.81 & 64.24 & 0.10 \\
& \texttt{Mistral-Small-3.2-24B-Instruct-2506} & 57.03 & 57.38 & 54.69 & 56.00 & 0.14 \\
& \texttt{Qwen3-14B} & 51.56 & 51.43 & 56.25 & 53.73 & 0.03 \\
& \texttt{Qwen3-235B-A22B-Instruct-2507} & 59.38 & 58.33 & 65.62 & 61.76 & 0.19 \\
& \textbf{Average} & \textbf{55.47} & \textbf{55.50} & \textbf{58.85} & \textbf{56.39} & \textbf{0.11} \\
\midrule

\multirow{7}{*}{RAP-BERT}
& \texttt{gemma-3-27b-it} & 49.22 & 49.32 & 56.25 & 52.55 & -0.02 \\
& \texttt{Meta-Llama-3.1-70B-Instruct} & 56.25 & 54.88 & 70.31 & 61.64 & 0.13 \\
& \texttt{Meta-Llama-3.1-8B-Instruct} & 53.12 & 51.79 & 90.62 & 65.91 & 0.09 \\
& \texttt{Mistral-Small-3.2-24B-Instruct-2506} & 56.25 & 55.41 & 64.06 & 59.42 & 0.13 \\
& \texttt{Qwen3-14B} & 55.47 & 54.79 & 62.50 & 58.39 & 0.11 \\
& \texttt{Qwen3-235B-A22B-Instruct-2507} & 64.06 & 59.18 & 90.62 & 71.60 & 0.33 \\
& \textbf{Average} & \textbf{55.73} & \textbf{54.23} & \textbf{72.40} & \textbf{61.59} & \textbf{0.13} \\
\midrule

\multirow{7}{*}{P-SFT}
& \texttt{qwen3-4b} & 88.28 & 87.69 & 89.06 & 88.37 & 0.77 \\
& \texttt{qwen3-8b} & 81.25 & 85.71 & 75.00 & 80.00 & 0.63 \\
& \texttt{qwen3-14b} & 78.91 & 78.46 & 79.69 & 79.07 & 0.58 \\
& \texttt{llama3.2-3b} & 76.56 & 82.69 & 67.19 & 74.14 & 0.54 \\
& \texttt{llama3.1-8b} & 85.16 & 92.45 & 76.56 & 83.76 & 0.71 \\
& \texttt{ministral-8b} & 93.75 & 96.67 & 90.62 & 93.55 & 0.88 \\
& \textbf{Average} & \textbf{83.98} & \textbf{87.28} & \textbf{79.69} & \textbf{83.15} & \textbf{0.68} \\
\bottomrule
\end{tabular}%
}
\caption{Task 2 results across different methods and models.}
\label{tab:task2_results}
\end{table*}

\onecolumn

\begin{tcolorbox}[title=Prompt for Persona Card Generation, left=2mm,right=1mm,top=0mm, bottom=0mm,colback=white]
    \begin{lstlisting}[style=plain]
Please create a realistic and lifelike virtual character profile based on the clues provided, which must include the following fields: Name, Age, Gender, Job Title, Job Responsibilities, Education Level, and Interests.

Generation Goals:
1. Ensure the generated profile aligns with everyday social common sense, maintaining logical consistency between the character's age, gender, education, and occupation.
2. Within this logical framework, include creative and individualized details to enhance realism and diversity.

Generation Requirements:
1. Name: Randomly select one name from the provided list as the character's name. Candidate names: {candidate_names_list}.
2. Occupation: Generate based on the following clues: this occupation is generally called {job_title}, and is also known as {reported_title_list}. The typical responsibilities of this job include {job_task_list}. Please determine the final specific job title and responsibilities accordingly. The 'Job Responsibilities' field must consist of two natural-language paragraphs, each describing a different work task or typical scenario. Each paragraph may show individual variation, such as area of expertise, working style, common collaborators, or daily rhythm.
3. Education Level: People in this occupation usually have one of the following educational backgrounds: {education_list}. Randomly select one that is logically consistent and matches both the character's age and occupation.
4. Interests: The character's interest type belongs to {riasec} from the Holland Code. {riasec_definition}. In an interest inventory test, the character selected the following activities: {riasec_scale}. Note that this inventory assumes the respondent imagines activities they would enjoy, rather than those they are currently skilled at or engaged in. Based on the above definition and selected activities, create two natural-language paragraphs describing two distinct personal interests of the character. Each interest should be specific and lifelike, different from the listed activities yet consistent with the {riasec} type, and completely unrelated to the character's occupation.
* All descriptions in 'Job Responsibilities' and 'Interests' must be written in the first person, as if the character is introducing themselves.

Please strictly follow the format below and output nothing else. For the 'Job Responsibilities' and 'Interests' fields, use (1) (2) to distinguish different paragraphs. Each field must occupy exactly one line, resulting in a total of seven lines in the output.

Name:
Age:
Gender:
Job Title:
Job Responsibilities: (1) (2)
Education Level:
Interests: (1) (2)
    \end{lstlisting}
    \end{tcolorbox}

\newpage
\begin{tcolorbox}[title=Prompt for Occupation Experience Generation, left=2mm,right=1mm,top=0mm, bottom=0mm,colback=white]
    \begin{lstlisting}[style=plain]
Please write a realistic and lifelike personal experience in the first-person perspective, based on the character profile I provide. You should comprehensively reference the character's age, gender, job content, education level, and Big Five personality traits when creating the text.

Generation Goals:
The generated content should align with social common sense and remain logically consistent with the character's profile. Within that logical framework, incorporate individualized and creative details to make the character vivid, believable, and diverse.

Generation Requirements:
1. Write a natural, authentic, and detailed personal experience centered on the character's job content and personality traits, about 150 words in length. The character's basic information is as follows: Age: {age}, Gender: {gender}, Education Level: {education_level}. You do not need to explicitly state these basic details in the personal experience.
2. The character's job content is as follows: {job_task}
3. The character's Big Five personality test result shows a {bigfive_value} score in the {bigfive} dimension. {bigfive_definition} Individuals with this personality tendency typically show the following behavioral characteristics: {bigfive_behavior_list}. Please select several of these behaviors, combine them with the definition of this personality dimension and the job content, and craft a credible personal experience.

Please output only the personal experience text, without any explanations or additional information.
    \end{lstlisting}
    \end{tcolorbox}

\begin{tcolorbox}[title=Prompt for Occupation-Related Dialogues Generation, left=2mm,right=1mm,top=0mm, bottom=0mm,colback=white]
    \begin{lstlisting}[style=plain]
Please create a natural and authentic conversation containing 5-8 exchanges. One exchange = one user turn followed by one chatbot turn.

Generation Requirements:
1. Chatbot Setting: The chatbot is a conversational partner. Its tone is gentle and curious. It listens attentively like a caring friend, naturally asks follow-up questions to encourage sharing, and maintains a natural, rhythmic flow in the dialogue.

2. User Setting: The user's basic information is as follows: Name: {name}, Age: {age}, Gender: {gender}, Education Level: {education_level}, and their Big Five personality dimension {bigfive} is very {bigfive_value}. The conversation does not need to intentionally prompt the user to state these details.

3. User Speaking Style: The following are examples of the user's typical way of speaking. These examples are not related to the content you will generate. Please only imitate the tone, wording, and style demonstrated in these examples when generating the dialogue: {user_speaking_style}.

4. Conversation Content: The dialogue should organically unfold the following work-related self-statement from the user: {job_experience}. The user should gradually reveal this information across multiple turns rather than stating it all at once. The chatbot should naturally guide the user to share their full story and mention at least once that the discussion appears to be work-related, using questions and responses to make the conversation authentic and vivid. By the end of the dialogue, the full meaning and key details of the above self-statement should have been expressed through the user's responses. You must also capture the characteristics in the self-statement that reflect the user's Big Five personality dimension {bigfive} being very {bigfive_value}, but without explicitly mentioning, discussing, or stating their personality traits. You must also capture the details that reveal the nature of the user's work tasks.

Output Format:
Use the following alternating format for each turn, where each line contains the speaker and their dialogue. Output only the full conversation text without any explanations or additional notes.
user:
chatbot:
    \end{lstlisting}
    \end{tcolorbox}

\newpage
\begin{tcolorbox}[title=Prompt for Interest Activity Experience Generation, left=2mm,right=1mm,top=0mm, bottom=0mm,colback=white]
    \begin{lstlisting}[style=plain]
Please write a realistic and lifelike description of the character's personal interest in the first-person perspective, based on the character profile I provide. You should comprehensively reference the character's age, gender, education level, and personal interest statement when creating the text.

Generation Goals:
The generated content should align with social common sense and remain logically consistent with the character's profile. Within that logical framework, incorporate individualized and creative details to make the character vivid, believable, and diverse.

Generation Requirements:
1. The character's basic information is as follows: Age: {age}, Gender: {gender}, Education Level: {education_level}. You do not need to explicitly state these basic details in the text.
2. Write a natural and authentic passage introducing the character's personal interest, about 150 words in length. The passage should consider that it is the character's personal hobby, entirely unrelated to their main occupation. The character's personal interest statement is as follows: {riasec_activity}

Please output only the passage introducing the personal interest, without any explanations or additional information.
    \end{lstlisting}
    \end{tcolorbox}

\begin{tcolorbox}[title=Prompt for Activity-Related Dialogues Generation, left=2mm,right=1mm,top=0mm, bottom=0mm,colback=white]
    \begin{lstlisting}[style=plain]
Please create a natural and authentic conversation containing 5-8 exchanges. One exchange = one user turn followed by one chatbot turn.

Generation Requirements:
1. Chatbot Setting: The chatbot is a conversational partner. Its tone is gentle and curious. It listens attentively like a caring friend, naturally asks follow-up questions to encourage sharing, and maintains a natural, rhythmic flow in the dialogue.

2. User Setting: The user's basic information is as follows: Name: {name}, Age: {age}, Gender: {gender}, Education Level: {education_level}, and their Holland Code is {riasec_value}. The conversation does not need to intentionally prompt the user to state these details.

3. User Speaking Style: The following are examples of the user's typical way of speaking. These examples are not related to the content you will generate. Please only imitate the tone, wording, and style demonstrated in these examples when generating the dialogue: {user_speaking_style}.

4. Conversation Content: The dialogue should organically unfold the following self-statement related to the user's personal interests and hobbies: {riasec_experience}. The user should gradually reveal this information across multiple turns rather than stating it all at once. The chatbot should naturally guide the user to share their interests outside of work and mention at least once that the discussion focuses on non-work hobbies, using questions and responses to make the conversation authentic and vivid. By the end of the dialogue, the full meaning and key details of the above self-statement should have been conveyed through the user's responses. You must capture the details that reveal the user's interests align with the Holland code category {riasec_value}, but without explicitly mentioning, discussing, or stating their Holland code.

Output Format:
Use the following alternating format for each turn, where each line contains the speaker and their dialogue. Output only the full conversation text without any explanations or additional notes.
user:
chatbot:
    \end{lstlisting}
    \end{tcolorbox}

\end{document}